\documentclass{article}

\usepackage{microtype}
\usepackage{graphicx}
\usepackage{subcaption}
\usepackage{booktabs}
\usepackage{hyperref}

\usepackage[preprint]{icml2026}

\usepackage{amsmath,amssymb,amsfonts}

\usepackage{multirow}
\usepackage{xcolor}
\usepackage{colortbl}
\usepackage{array}
\usepackage{enumitem}
\usepackage{fontawesome5}

\definecolor{improvement}{RGB}{34, 139, 34}
\definecolor{decline}{RGB}{178, 34, 34}
\definecolor{highlight}{RGB}{255, 248, 220}

\newcommand{\up}[1]{\textcolor{improvement}{#1}}
\newcommand{\down}[1]{\textcolor{decline}{#1}}

\icmltitlerunning{Scaling Multiagent Systems with Process Rewards}

\begin{document}

\twocolumn[
  \icmltitle{Scaling Multiagent Systems with Process Rewards}

  \icmlsetsymbol{equal}{*}

  \begin{icmlauthorlist}
    \icmlauthor{Ed Li}{equal,yale}
    \icmlauthor{Junyu Ren}{equal,uchicago}
    \icmlauthor{Cat Yan}{oxford}
  \end{icmlauthorlist}

  \icmlaffiliation{yale}{Yale University}
  \icmlaffiliation{uchicago}{University of Chicago}
  \icmlaffiliation{oxford}{University of Oxford}

  \icmlcorrespondingauthor{Ed Li}{ed.li@yale.edu}

  \icmlkeywords{Multiagent Systems, Agentic Systems, Post-training, Reinforcement Learning, Large Language Models, LLM Agents, Process Rewards, LLM-as-a-Coach, Distributed Training}

  \vskip 0.3in
]

\printAffiliationsAndNotice{\icmlEqualContribution}

% Links
\begin{center}
\large
\href{https://github.com/ltjed/multiagent-coaching}{\faGithub~Code} \quad
\href{https://ltjed.github.io/MAPPA/}{\faNewspaper~Blog}
\end{center}
\vskip 0.2in

%==============================================================================
\begin{abstract}
%==============================================================================
While multiagent systems have shown promise for tackling complex tasks via specialization, finetuning multiple agents simultaneously faces two key challenges: (1) credit assignment across agents, and (2) sample efficiency of expensive multiagent rollouts. In this work, we propose finetuning multiagent systems with per-action process rewards from AI feedback (MAPPA) to address both. Through assigning credit to individual agent actions rather than only at task completion, MAPPA enables fine-grained supervision without ground truth labels while extracting maximal training signal from each rollout.
We demonstrate our approach on competition math problems and tool-augmented data analysis tasks. On unseen math problems, MAPPA achieves +5.0--17.5pp on AIME and +7.8--17.2pp on AMC. For data analysis tasks, our method improves success rate by +16.7pp while quality metrics improve by up to 47\%, validating that per-action supervision can lead to improvements across different multiagent systems on various domains. By addressing these challenges, our work takes a first step toward scaling multiagent systems for complex, long-horizon tasks with minimal human supervision.
\end{abstract}

\section{Introduction}
\begin{figure}[t]
  \centering
  \includegraphics[width=\columnwidth]{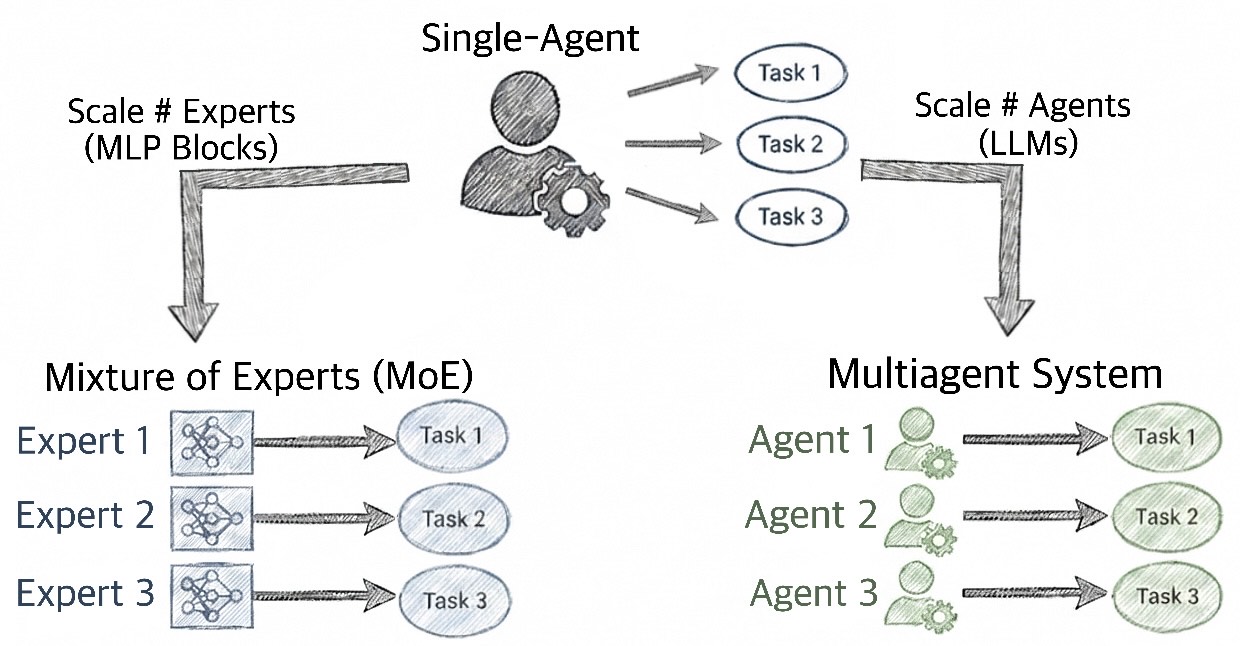}
  \caption{Multiagent architectures with separate weights enable specialization through end-to-end training, sidestepping catastrophic forgetting that limits single-model scaling just like mixture-of-experts (MoE) architecture.}
  \label{fig:agent_scaling}
  \end{figure}
Large reasoning models trained with reinforcement learning have emerged as the dominant paradigm for state-of-the-art AI systems~\citep{openai2024o1, deepseek-r1-2025, scaling-test-time-compute-2025}. Augmented with external tools such as code interpreters and search engines~\citep{schick2023toolformer}, these models achieve remarkable performance on mathematical reasoning~\citep{shao2024deepseekmathpushinglimitsmathematical}, coding~\citep{jimenez2024swebench}, and scientific problems~\citep{novikov2025alphaevolve}. Building on these capabilities, multiagent systems---where multiple language model-based agents interact to solve complex tasks---offer an intuitive path forward as specializations and the diversity of perspectives can tackle problems better than any single agent.~\citep{wu2023autogen, metagpt}.

The bulk of existing multiagent work implements specialization through prompt engineering: assigning different system prompts, personas, or tool access to agents within the system. While effective for orchestrating pre-trained models, the true strength of multiagent systems lies in the ability to \emph{modify the weights} of individual agents, enabling specialization without interference. Recent evidence suggests such specialization naturally emerges through reinforcement learning: reasoning models trained solely for accuracy spontaneously develop diverse internal personas and expertise~\citep{kim2026societies}. By separating weights across agents, improving one agent's capabilities cannot degrade another's performance, allowing scaling without catastrophic forgetting (Figure~\ref{fig:agent_scaling}).

Recent work has begun exploring this direction of fine-tuning agents in multiagent systems simultaneously~\citep{subramaniam2025multiagent, liao2025marft, park2025maporl, liu2025magrpo}. However, two significant challenges remain:
\begin{enumerate}
    \item \textbf{Credit assignment}: How do we attribute the performance of the overall system to actions taken by individual agents?
    \item \textbf{Computational cost}: Rollouts of the entire multiagent system can take minutes or even hours, yet yield only a single outcome reward signal.
\end{enumerate}

We address these challenges through finetuning \textbf{M}ulti\textbf{A}gent system with \textbf{P}er-action \textbf{P}rocess rewards from \textbf{A}I feedback (\textbf{MAPPA}). Rather than assigning a single sparse reward at trajectory's end according to fixed instructions in LLM-as-a-Judge, we leverage language models as \emph{coaches} to assess the quality of each agent's action given its role, inputs, and environment feedback such as tool execution results. This yields dense learning signal throughout the trajectory, enabling effective training even when tasks fail entirely. Crucially, the coach performs implicit credit assignment: when a downstream agent encounters a file-not-found error for an artifact that an upstream agent should have produced, the coach assigns low scores to the upstream agent's faulty actions rather than penalizing the downstream agent. Furthermore, the number of training signals now scales with the number of actions taken, dramatically improving sample efficiency over outcome-based methods.

We validate this approach on two separate domains: mathematical reasoning with code execution (MathChat) and end-to-end data science pipelines (DSBench~\citep{jing2025dsbench}). On MathChat, our three-agent system achieves improvements of \textbf{+5.0--17.5pp on AIME} and \textbf{+7.8--17.2pp on AMC} across two model configurations. On DSBench, where multiagent pipelines must engineer features, train models, and generate predictions, training improves success rate by \textbf{+16.7pp} while quality metrics improve by up to \textbf{47\%}.

We present a general framework for training multiagent systems on complex, tool-augmented tasks via coach-guided reinforcement learning. The key components---agent topology, reward structure, and training pipeline---are domain-agnostic and can be configured to support diverse applications. Our source code is available at \url{https://github.com/ltjed/multiagent-coaching}. Overall, our results suggest that \textbf{scaling the number of specialized agents represents a promising new dimension for improving performance on complex, long-horizon tasks}.

%==============================================================================
\section{Methodology}
\label{sec:method}
%==============================================================================

%------------------------------------------------------------------------------
\subsection{Problem Setting}
\label{sec:framework}
%------------------------------------------------------------------------------
MAPPA finetunes tool-augmented multiagent systems where LLM agents collaborate to solve tasks. Each agent is initialized from a pretrained LLM with independent policy parameters trained separately, enabling specialization. All agents live in a terminal environment and execute in a pre-defined topology (e.g., sequential pipelines, debate, mixture-of-agents). Within each agent's turn, it generates one or more actions that may include tool calls. When tool calls are generated as part of an agent's action, they are processed and executed in a sandboxed environment, the agent will then receive environment feedback (e.g., stdout, stderr, error messages). Depending on the task, the agent may also need to call tools to read, write files or other artifacts in the terminal environment.
%
%------------------------------------------------------------------------------
\subsection{Process Rewards from AI Feedback}
\label{sec:coach}
%------------------------------------------------------------------------------
Multi-turn tool use with inter-agent dependencies creates challenging credit assignment---which agent's actions led to success or failure? Unlike existing RL approaches that rely on sparse outcome rewards (success/failure for entire trajectories), we leverage a \emph{coach} LLM to perform per-action credit assignment, providing dense process-quality rewards on a 0--10 scale.\footnote{We found the 0--10 scale empirically outperforms continuous 0--1 scores, likely due to the prevalence of 0--10 rating samples in the pre-training corpus, i.e.,  a ``$7$'' or a ``$4$,'' carries more semantic meaning than decimal values like $0.305$.} We use the term ``coach'' rather than ``judge'' to emphasize that while LLM-as-a-judge evaluates final outputs against fixed rubrics, a coach provides context-aware feedback on each intermediate action to guide improvement throughout the trajectory. The coach evaluates each action holistically based on: (1) the agent's role and responsibilities, (2) the input context the agent observed, (3) the agent's action, and (4) tool execution results (stdout, stderr, error messages) when the output contains tool calls. This enables the coach to identify the responsible agent when failures occur: if a downstream agent encounters a missing file error, the coach assigns low scores to the upstream agent that should have produced it, not the downstream agent that correctly reported the issue. See Appendix~\ref{app:coach_examples} for example coach evaluations demonstrating this context-aware feedback.

Crucially, MAPPA can work \textit{with or without ground truth/verifier} that most other alternatives RL approaches such as RLVR~\citep{lightman2023lets,deepseek-r1-2025} require. When ground truth is available (e.g., ROC-AUC, F1, RMSE for DSBench), the coach incorporates it to inform process scores via intelligently synthesizing multi-dimensional metrics into a single reward; When ground truth is unavailable, the coach simply evaluates how sensible it is for the agent to take that action given its context to the best of the coach's ability. In this work, ground truth is only provided for the last action of the last agent in the multiagent system.

\begin{figure}[ht]
\centering
\includegraphics[width=\columnwidth]{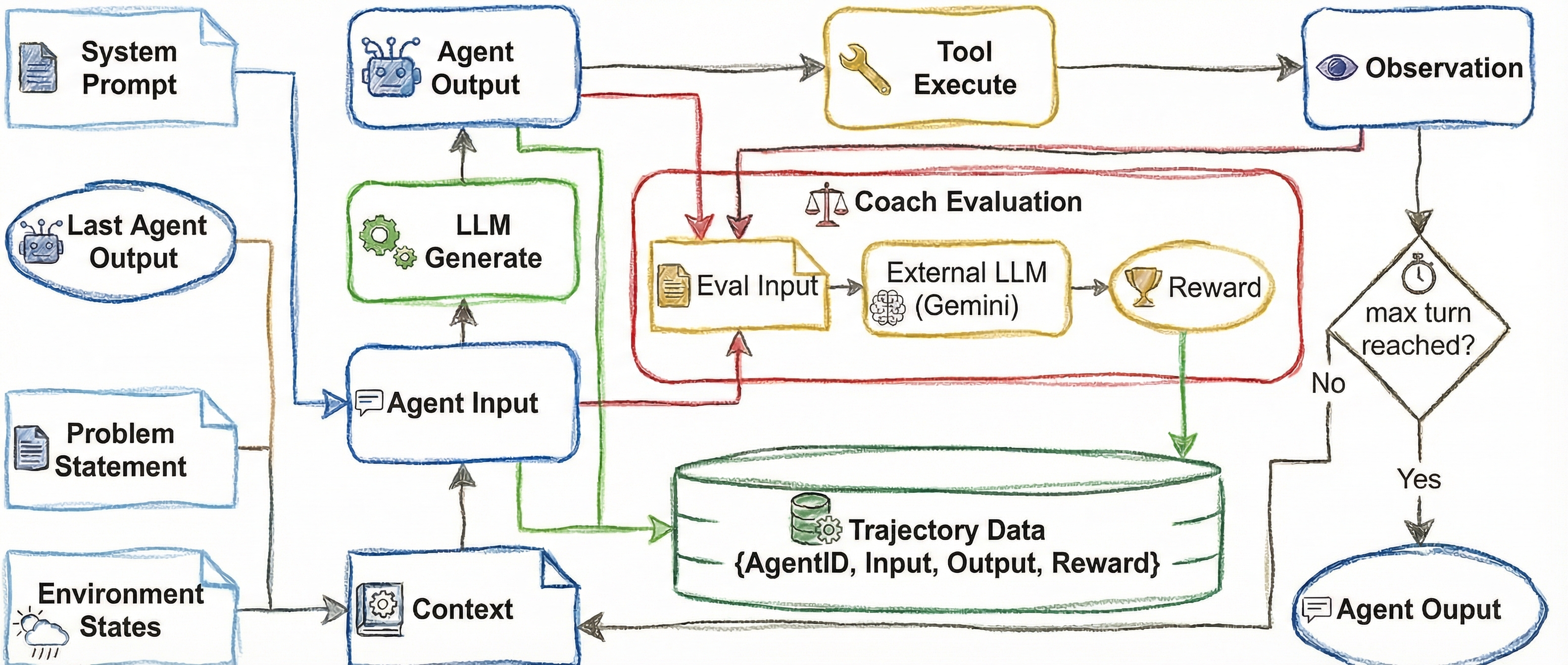}
\caption{Agent execution loop with per-action coach evaluation.}
\label{fig:agent_coach_flow}
\end{figure}

%------------------------------------------------------------------------------
\subsection{Training Algorithm}
\label{sec:training}
%------------------------------------------------------------------------------
As shown in Figure~\ref{fig:agent_coach_flow}, each agent action generates a trajectory tuple $(\text{agent\_id}, \text{input}, \text{action}, \text{reward})$ where the reward is provided by the coach evaluation. We then finetune each agent's underlying model with its actions' corresponding tuples via REINFORCE++~\citep{hu2025reinforceplusplus}. We choose REINFORCE++ rather than Group-Relative Policy Optimization (GRPO)~\citep{deepseek-r1-2025} for a fundamental reason: GRPO normalizes advantages within groups of samples sharing the same prompt, assuming identical input states. This assumption breaks in end-to-end multiagent training---each agent's input depends on upstream agents' stochastic outputs, so even rollouts from the same initial prompt produce different intermediate states. REINFORCE++ instead applies global batch normalization across all experiences, naturally handling state diversity. Global normalization also reduces variance from noisy coach rewards compared to deterministic ground truth metrics.
Each action receives an independent coach reward $r_t^{\text{coach}} \in [0, 1]$, with KL penalties computed per-token. The KL-penalized reward and advantage are:
\begin{equation}
r_t = r_t^{\text{coach}} - \beta \cdot D_{\text{KL}}(\pi_\theta \| \pi_{\text{ref}}), \quad A_t = \sum_{\tau \geq t} r_\tau
\end{equation}
where $\beta = 0.01$ and $A_t$ is the undiscounted return-to-go, propagating downstream rewards to earlier actions. Advantages are then globally normalized across all agents and experiences in the batch: $\hat{A}_t = (A_t - \mu) / \sigma$. The policy gradient uses the standard PPO clipped objective with $\epsilon = 0.2$. See Appendix~\ref{app:training} for full derivations.
%
%------------------------------------------------------------------------------
\subsection{Scalable Distributed Training Architecture}
\label{sec:implementation}
%------------------------------------------------------------------------------
On-policy RL requires fresh trajectories from the current policy for each gradient update, creating an efficiency challenge for multiagent systems where rollouts are expensive. Our training infrastructure, extended from MARTI~\citep{marti2025}, addresses this through tight coupling: each iteration consists of (1) parallel trajectory collection, (2) coach evaluation and experience preparation, and (3) synchronous gradient updates---then immediately proceeds to the next iteration. Training uses Ray for distributed coordination, vLLM for inference, and DeepSpeed ZeRO-3 for memory-efficient updates.
Each agent maintains independent actor groups initialized in parallel: vLLM engines for inference, reference models for KL computation, and policy actors for training, with optional model co-location on shared GPUs to reduce memory overhead. During rollout, prompts are sharded across workers that execute multiagent workflows independently, with coach evaluation overlapped asynchronously to reduce wall-clock time. Completed trajectories are routed to originating agents for gradient computation; NCCL broadcasts updated weights to vLLM engines after each training step. All agents share the training loop but maintain independent policy parameters. See Appendix~\ref{app:implementation} for details.

%==============================================================================
\section{Results}
\label{sec:results}
%==============================================================================
We demonstrate our approach on two separate multiagent systems across two domains: competition math with code execution (MathChat) and end-to-end data science pipelines (DSBench). While competition math can be solved by single agents~\citep{openai2024o1,deepseek-r1-2025}, applying MAPPA to multiagent systems with different underlying models and constraints tests the generality and robustness. DSBench, in contrast, more closely resembles real-world complex tasks that multiagent systems are built for.
%
%------------------------------------------------------------------------------
\subsection{MathChat: Competition Math with Code Execution}
\label{sec:exp_mathchat}
%------------------------------------------------------------------------------
\paragraph{Task and Dataset.}
MathChat involves solving AIME (American Invitational Mathematics Examination) competition math problems with code execution. We randomly sample 512 problems from 1983--2024 for training and use disjoint held-out sets for evaluation: 30 AIME problems from 2025 and 32 AMC (American Mathematics Competition, similar to but slightly simpler than AIME) problems.

\paragraph{Multiagent Configuration.}
We implement a three-agent sequential pipeline designed to separate reasoning, computation, and verification:
\begin{itemize}[nosep]
    \item \emph{Problem Solver}: Drafts step-by-step reasoning for downstream agents. The prompt instructs: ``Your job is to draft a solution to the problem...''
    \item \emph{Code Executor}: Has the option to write and execute Python code to verify and compute solutions. The prompt instructs: ``You can execute Python code. Write code in \texttt{```python```} blocks and it will be automatically executed.''
    \item \emph{Verifier}: Synthesizes information from upstream agents and outputs the final answer. The prompt instructs: ``You are the last agent. The system succeeds only if YOU output the correct answer.'' The final answer must be formatted as \texttt{\textbackslash boxed\{answer\}} to be counted as correct.
\end{itemize}
Each agent is initialized from the same pretrained checkpoint but finetuned separately, enabling specialization. We experiment with two reasoning models: DeepSeek-R1-Distill-Qwen-1.5B~\citep{deepseek-r1-2025} and Qwen3-4B~\citep{qwen2025}. All agents use reasoning model inference with \texttt{<think>} tags, allowing internal deliberation before generating outputs. Each agent has a 4K token output limit after which the response is cut off, content within \texttt{<think>} tags included. See Appendix~\ref{app:mathchat_prompts} for complete agent prompts.

\paragraph{Training Configuration.}
We finetune the multiagent systems using REINFORCE++ with Gemini 2.5 Flash as the coach on 8 NVIDIA H100 GPUs. Key hyperparameters: actor learning rate $10^{-6}$, rollout batch size 32, training temperature 1.0, evaluation temperature 0.6, KL penalty coefficient $\beta=0.01$. With 512 training problems processed in rollout batches of 32, one epoch equals 16 training steps (one complete pass through all problems).
% Different runs train for varying durations: the DeepSeek run completes 106 steps ($\sim$6.6 epochs), while Qwen runs train longer. 
We evaluate every 4 steps (every 0.25 epochs) on held-out problems. Each evaluation samples every problem 4 times and reports mean accuracy to reduce variance from stochastic generation. See Appendix~\ref{app:computational_cost} for computational cost details.

\begin{table}[t]
\centering
\caption{MathChat performance on held-out competition math problems.}
\label{tab:mathchat_performance}
\small
\begin{tabular}{@{}lcc@{}}
\toprule
& R1-Distill-Qwen-1.5B & Qwen3-4B \\
\midrule
\textbf{AMC} & & \\
\quad Baseline & 60.9\% & 78.1\% \\
\quad Best & 78.1\% & 85.9\% \\
\quad $\Delta$ & \up{+17.2pp} & \up{+7.8pp} \\
\midrule
\textbf{AIME} & & \\
\quad Baseline & 24.2\% & 49.2\% \\
\quad Best & 29.2\% & 66.7\% \\
\quad $\Delta$ & \up{+5.0pp} & \up{+17.5pp} \\
\bottomrule
\end{tabular}
\end{table}

\paragraph{Results.}
As shown in Table~\ref{tab:mathchat_performance}, MAPPA improves performance across both model configurations, with gains ranging from +5.0pp to +17.5pp depending on model capacity and task difficulty. The larger Qwen3-4B achieves the biggest improvement (+17.5pp) on AIME, while the smaller DeepSeek-R1-Distill-Qwen-1.5B shows asymmetric gains: +17.2pp on the more accessible AMC but only +5.0pp on AIME 2025, suggesting capacity limitations on harder reasoning tasks.

\begin{figure}[t]
\centering
\includegraphics[width=\columnwidth]{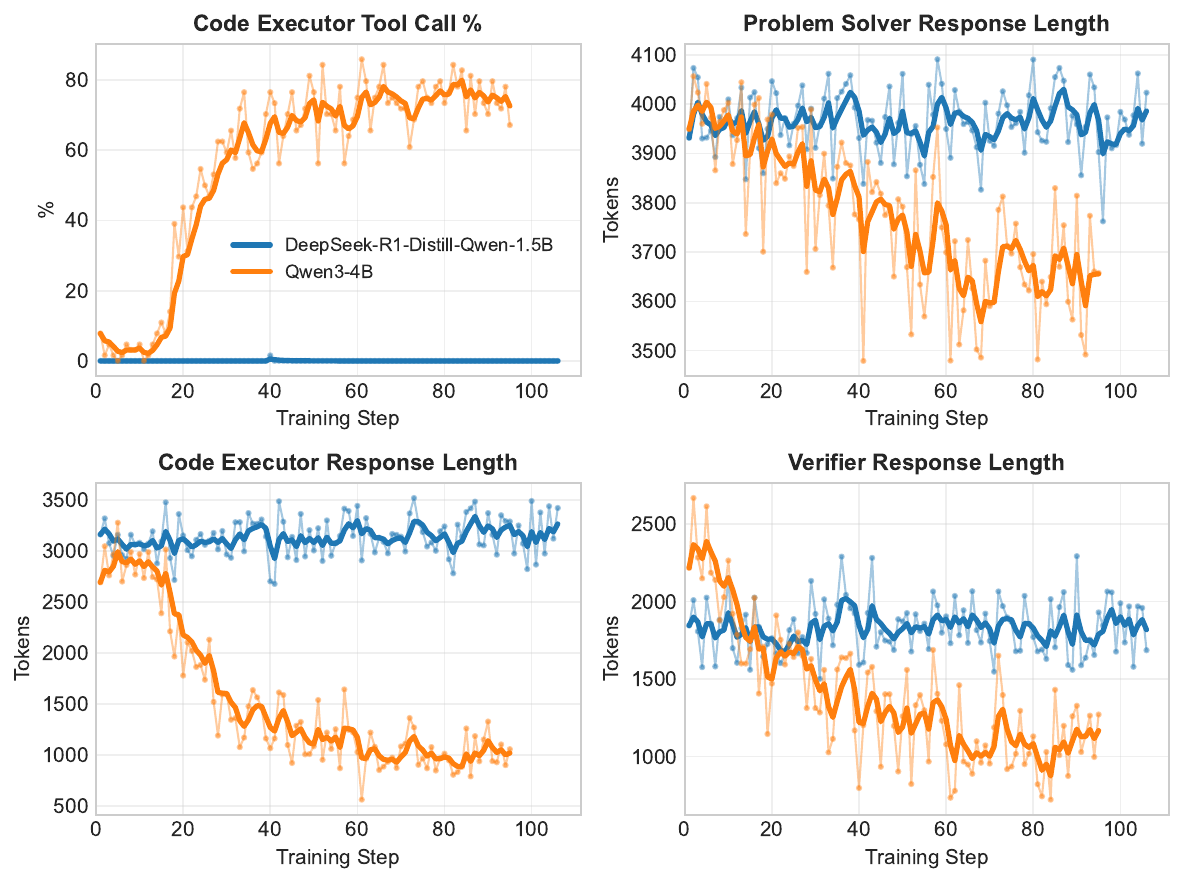}
\caption{Behavioral metrics during MathChat training. Code Executor Tool Call \% (top-left) shows the percentage of Code Executor actions that contain a tool call; the other three panels show average response length (in tokens) for each agent at each training step. Faint lines show raw data; solid lines show exponential moving average ($\alpha=1/3$). Qwen3-4B shows dramatically increased tool usage and more concise responses, while R1-Distill-Qwen-1.5B maintains stable behavior throughout training.}
\label{fig:mathchat_behavioral_main}
\end{figure}

Interestingly, behavioral metrics reveal divergent learning patterns between the two models (Figure~\ref{fig:mathchat_behavioral_main}). The larger Qwen3-4B shows dramatic behavioral adaptation: successful tool call rate increases substantially while response lengths decrease across all three agents. In contrast, the smaller 1.5B model maintains relatively stable behavioral metrics throughout training. This suggests that with greater capacity, Qwen3-4B learns to leverage tool calls more effectively through MAPPA, while the 1.5B model improves accuracy without developing such qualitative behavioral changes---demonstrating that process rewards can drive improvement even when model capacity limits behavioral adaptation. We also evaluate under partial information constraints, where each agent observes only the immediately preceding agent's output with no access to earlier context; results in Appendix~\ref{app:mathchat_partial} show MAPPA still achieves consistent improvements (+3.9--5.8pp), further demonstrating robustness.

%------------------------------------------------------------------------------
\subsection{DSBench: Data Science Pipelines}
\label{sec:exp_dsbench}
%------------------------------------------------------------------------------

\paragraph{Task and Dataset.}
DSBench presents 72 Kaggle-style machine learning tasks requiring complete end-to-end pipeline execution. Each task provides training/test CSV files, task type (classification/regression), and ground truth labels for evaluation. We use 64 tasks for training and 6 for held-out evaluation (4 classification, 2 regression)---an 8.6\% held-out ratio that is reasonable given the limited size of existing data science benchmarks.
\paragraph{Multiagent Configuration.}
We implement a three-agent sequential pipeline where each agent specializes in a distinct phase: the \emph{Data Engineer} performs exploratory data analysis, preprocessing, and feature engineering (maximum 3 turns); the \emph{Modeler} handles algorithm selection, model training, and hyperparameter tuning (maximum 5 turns); and the \emph{Analyst} generates final predictions and formats submissions (maximum 2 turns). Each agent is initialized from the same pretrained checkpoint (Qwen3-4B) with independent policy parameters.
Agents communicate via base64-encoded file passing: the Data Engineer produces \texttt{X\_train.pkl}, \texttt{y\_train.pkl}, \texttt{X\_test.pkl}; the Modeler consumes these and produces \texttt{model.pkl}; the Analyst loads both to generate \texttt{submission.csv}. This explicit file dependency enables the coach to assign blame when pipeline failures occur. Figure~\ref{fig:dsbench_pipeline} illustrates this three-agent pipeline; see Appendix~\ref{app:dsbench_prompts} for complete agent prompts.

\begin{figure*}[t]
\centering
\includegraphics[width=\textwidth]{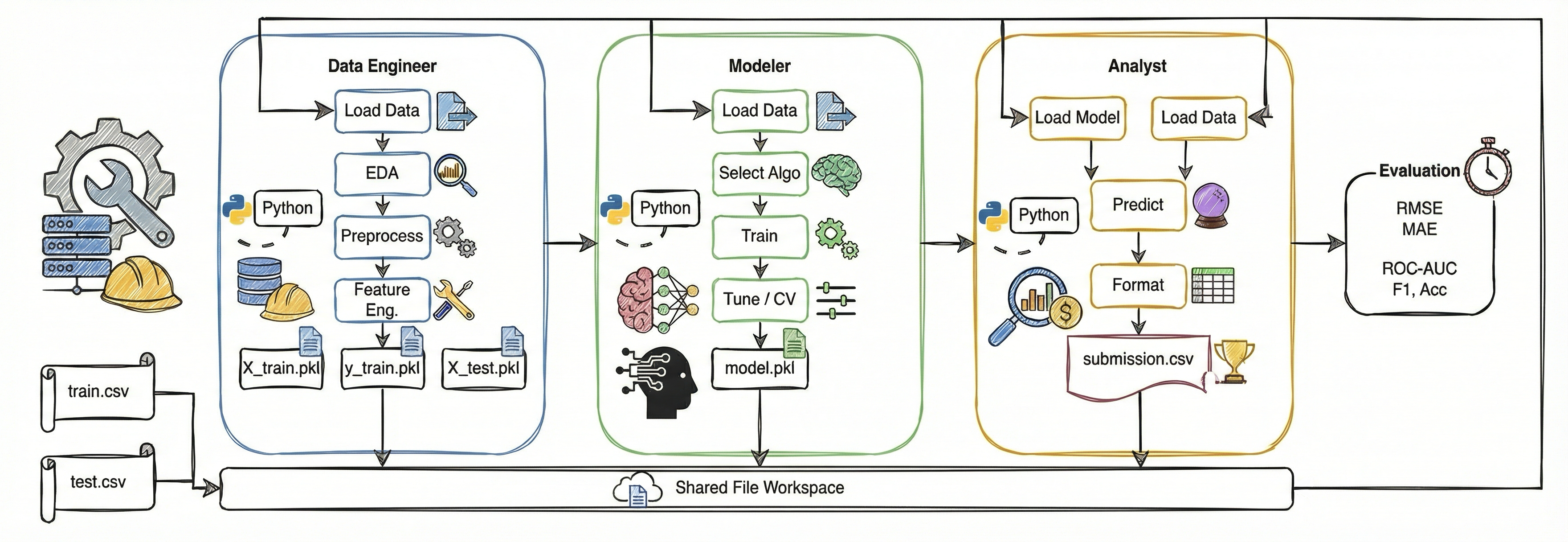}
\caption{DSBench three-agent pipeline. Each agent executes Python code via a shared sandbox, reading inputs from and writing outputs to a shared file workspace. The Data Engineer preprocesses raw CSV files into pickle artifacts; the Modeler trains and saves a model; the Analyst generates the final submission for evaluation.}
\label{fig:dsbench_pipeline}
\end{figure*}

\paragraph{Training Configuration.}
We train using REINFORCE++ with Gemini 2.5 Pro as the coach. All agents have access to SandboxFusion for Python code execution. When ground truth labels are available, we compute task-specific metrics (Accuracy, F1 for classification; RMSE, MAE for regression) and provide them to the coach. Rather than naively averaging these metrics or picking a single one, the coach intelligently weighs them in context---penalizing reward hacking (e.g., optimizing one metric at the expense of others), attributing poor metrics to the responsible agent (Data Engineer's preprocessing vs.\ Analyst's predictions), and producing a fair process score that reflects overall execution quality. Training uses Ray, vLLM, and DeepSpeed ZeRO-3 on 8 NVIDIA H100 GPUs. Key hyperparameters: actor learning rate $10^{-6}$, rollout batch size 16, training temperature 1.0, evaluation temperature 0.6. Checkpoints are evaluated every 2 steps on 6 held-out tasks.

\begin{figure}[t]
\centering
\includegraphics[width=\columnwidth]{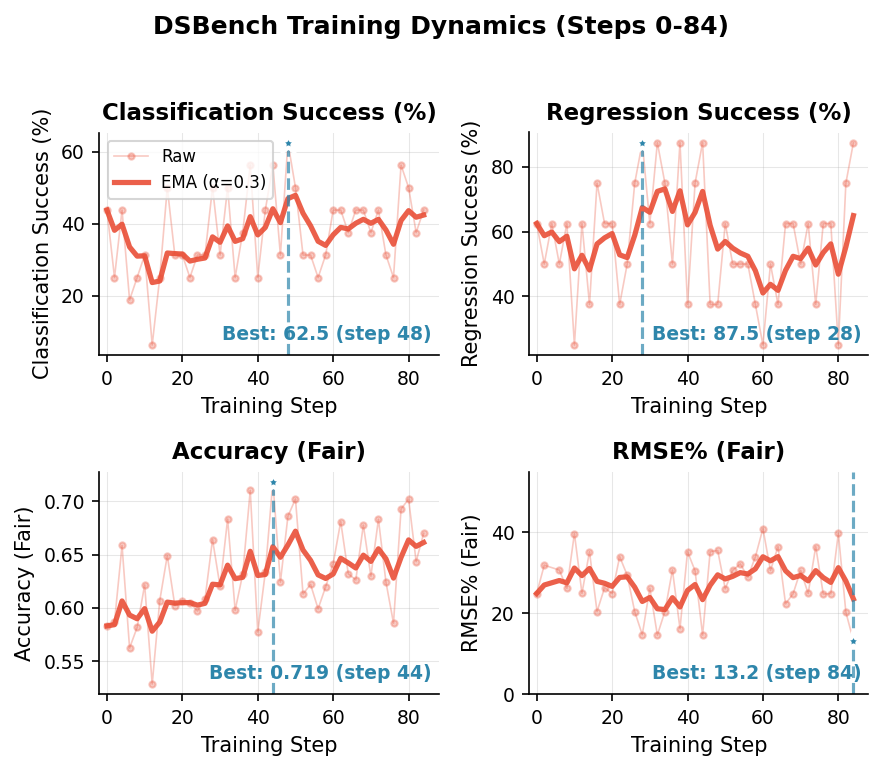}
\caption{DSBench training dynamics over 84 steps (21 epochs). Light points show raw metrics; solid line shows EMA ($\alpha$=0.3). Dashed lines mark peak raw values. Classification metrics peak early (steps 44--48) then decline, while regression RMSE continues improving through step 84, illustrating specialization to regression tasks.}
\label{fig:dsbench_dynamics}
\end{figure}

\paragraph{Results.}
The pipeline structure creates natural dependencies: the Modeler cannot train without properly formatted data from the Data Engineer, and the Analyst cannot predict without the trained model from the Modeler. This makes DSBench an ideal testbed for evaluating multiagent credit assignment and process-focused training. We train for 21 epochs (84 steps) and evaluate checkpoints every 2 steps. Table~\ref{tab:dsbench_combined} presents metrics at three checkpoints: baseline (epoch 0), peak success rate (epoch 11), and late training (epoch 21). Figure~\ref{fig:dsbench_dynamics} shows the full training dynamics across all 84 steps.

\begin{table*}[t]
\centering
\caption{DSBench evaluation comparing baseline (epoch 0), peak success rate (epoch 11), and late training (epoch 21). Epoch 11 maximizes success rate and fair metrics across both task types. By epoch 21, the model specializes in regression: RMSE continues improving while classification metrics regress to baseline. Fair metrics penalize failures (0.5 for Accuracy, 0 for F1, 50\% for MAE/RMSE). Each epoch consists of 4 training steps.}
\label{tab:dsbench_combined}
\small
\begin{tabular*}{\textwidth}{@{\extracolsep{\fill}}lccccc@{}}
\toprule
Metric & Epoch 0 & Epoch 11 & $\Delta_{0 \to 11}$ & Epoch 21 & $\Delta_{11 \to 21}$ \\
\midrule
\multicolumn{6}{l}{\textit{Success Rate $\uparrow$ (24 eval samples: 16 classification + 8 regression)}} \\
\quad Classification & 43.8\% & \textbf{56.2\%} & \up{+12.4pp} & 43.8\% & \down{$-$12.4pp} \\
\quad Regression & 62.5\% & \textbf{87.5\%} & \up{+25.0pp} & \textbf{87.5\%} & 0pp \\
\quad Total & 50.0\% & \textbf{66.7\%} & \up{+16.7pp} & 58.3\% & \down{$-$8.4pp} \\
\midrule
\multicolumn{6}{l}{\textit{Classification Quality}} \\
\quad Accuracy $\uparrow$ (Raw) & 0.690 & \textbf{0.889} & \up{+28.8\%} & \textbf{0.889} & 0\% \\
\quad Accuracy $\uparrow$ (Fair) & 0.583 & \textbf{0.719} & \up{+23.3\%} & 0.670 & \down{$-$6.8\%} \\
\quad F1 $\uparrow$ (Raw) & 0.288 & \textbf{0.309} & \up{+7.3\%} & 0.263 & \down{$-$14.9\%} \\
\quad F1 $\uparrow$ (Fair) & 0.126 & \textbf{0.174} & \up{+38.1\%} & 0.115 & \down{$-$33.9\%} \\
\midrule
\multicolumn{6}{l}{\textit{Regression Quality}} \\
\quad MAE $\downarrow$ (Raw) & 7.1\% & 6.9\% & \up{$-$2.8\%} & \textbf{5.7\%} & \up{$-$17.4\%} \\
\quad MAE $\downarrow$ (Fair) & 23.2\% & 12.3\% & \up{$-$47.0\%} & \textbf{11.2\%} & \up{$-$8.9\%} \\
\quad RMSE $\downarrow$ (Raw) & 9.8\% & 9.5\% & \up{$-$3.1\%} & \textbf{8.0\%} & \up{$-$15.8\%} \\
\quad RMSE $\downarrow$ (Fair) & 24.9\% & 14.6\% & \up{$-$41.4\%} & \textbf{13.2\%} & \up{$-$9.6\%} \\
\midrule
\multicolumn{6}{l}{\textit{Coach Scores by Agent $\uparrow$ (0--10 scale)}} \\
\quad Data Engineer & 5.43 & 5.09 & \down{$-$6.3\%} & \textbf{5.60} & \up{+10.0\%} \\
\quad Modeler & 4.55 & \textbf{5.28} & \up{+16.0\%} & 5.10 & \down{$-$3.4\%} \\
\quad Analyst & 5.56 & \textbf{7.24} & \up{+30.2\%} & 7.17 & \down{$-$1.0\%} \\
\bottomrule
\end{tabular*}
\end{table*}

Training initially improves both success rate (+16.7pp) and quality across task types (Accuracy +28.8\%, RMSE $-$41.4\% fair), validating that coach rewards translate to downstream metric improvements. Extended training (epoch 11$\to$21) shows a pattern that might appear as overfitting: classification success drops from 56.2\% back to baseline 43.8\% ($-$12.4pp), and classification quality metrics decline (Accuracy-Fair $-$6.8\%, F1-Fair $-$33.9\%). However, regression tells a different story---success maintains its peak (87.5\%) while quality continues improving: RMSE drops from 9.5\% to 8.0\% ($-$15.8\%). This asymmetry suggests specialization to regression rather than simple overfitting; we investigate this phenomenon in Section~\ref{sec:specialization}. Fair metrics enable reliable model selection by identifying epoch 11 as optimal for balanced performance across both task types. The per-agent coach scores in Table~\ref{tab:dsbench_combined} demonstrate that credit assignment enables targeted learning, with each agent receiving feedback specific to its role rather than shared outcome-based rewards.

%==============================================================================
\section{Discussion}
%==============================================================================

Despite the potential of multiagent systems, recent work studying inference-time coordination finds that naively scaling agent count \emph{without changing weights} actually \emph{degrades} performance on sequential reasoning tasks by 39--70\%~\citep{kim2025scalingagents}. Our results show that finetuning multiagent systems with process rewards achieves substantial improvements on both MathChat and DSBench.

In this section, we first investigate an unexpected specialization pattern in DSBench where training improves regression performance while classification regresses, tracing this to systematic biases in coach scoring. We then discuss coach model selection and compare our approach with prior multiagent finetuning work. Finally, we outline limitations and future directions including trainable coaches and reward backpropagation.

\subsection{Investigating Specialization}
\label{sec:specialization}

The DSBench results reveal an unexpected pattern: while regression performance continues improving, classification metrics regress to baseline. To understand this specialization, we stratify coach scores by task type (Table~\ref{tab:coach_bias}). The analysis reveals a systematic pattern: the coach assigns higher scores to regression tasks across all agents, with deltas ranging from $+$0.51 to $+$1.80. Notably, the Data Engineer's delta widens from $+$1.15 to $+$1.67 over training, while the Analyst's narrows from $+$1.80 to $+$0.77. We hypothesize that agents learn to exploit the coach's preference for regression tasks, leading to the observed specialization. This pattern illustrates a fundamental limitation of stateless evaluation: our coach judges each action in isolation without awareness that its scores drive gradient updates, and thus cannot detect or correct emergent imbalances in its own scoring behavior. We discuss how future coaches might overcome this limitation in Section~\ref{sec:agent_coach}.

\begin{table}[h]
\centering
\small
\caption{Coach score delta (Regression $-$ Classification) by agent across training epochs. Positive values indicate higher scores for regression tasks. The Data Engineer's widening delta correlates with the observed specialization.}
\label{tab:coach_bias}
\begin{tabular}{lccc}
\toprule
Agent & Epoch 0 & Epoch 11 & Epoch 21 \\
\midrule
Data Engineer & +1.15 & +1.40 & +1.67 \\
Modeler & +0.51 & +1.06 & +1.02 \\
Analyst & +1.80 & +1.17 & +0.77 \\
\bottomrule
\end{tabular}
\end{table}

\subsection{Coach Model Choice}

Effective credit assignment in multiagent systems requires more than observing actions---it requires reasoning about causality. When a DSBench pipeline fails to produce predictions, the coach must determine whether the Data Engineer failed to save required files, the Modeler chose an inappropriate algorithm, or the Analyst made errors in the final processing stage. This root-cause analysis demands strong logical reasoning capabilities, substantial context windows to process full agent transcripts, and the ability to interpret tool outputs (code execution results, file operations, error messages). Additionally, the coach's information asymmetry---access to environment feedback invisible to agents---enables credit assignment that would be impossible from trajectory data alone.

We expect weaker models to also function as coaches, given two fundamental asymmetries: (1) \emph{information asymmetry}---the coach observes tool outputs and environment feedback that agents cannot see, and (2) \emph{task asymmetry}---critiquing an action upon reflection is easier than proposing one under uncertainty. The advantage of stronger coaches is primarily \emph{quantitative} rather than qualitative: frontier models like GPT-4, Claude-3.5-Sonnet, and Gemini-2.5 assign appropriate rewards in more situations with higher reliability, reducing noise in the training gradient signal and making training more sample- and epoch-efficient. Weaker coaches would still provide directionally correct feedback, but with higher variance.

This design also aligns with our perspective that scaling specialized agents represents a new dimension for improving performance (Figure~\ref{fig:agent_scaling}). Strong general-purpose models serve as coaches to fine-tune a ``swarm'' of smaller, specialized agents. Once trained, the multiagent system can collectively outperform the single strong model in both efficiency and performance ceiling: smaller models are cheaper to run in parallel, and specialized agents can develop domain expertise that generalist models lack. The coach bootstraps capabilities it cannot itself achieve through direct inference.

\subsection{Comparison with Prior Work}

Most existing multiagent frameworks like AutoGen~\citep{wu2023autogen}, MetaGPT~\citep{metagpt}, and CAMEL~\citep{li2023camel} focus on collaboration through prompt engineering but do not update agent weights. Recently, more work has begun experimenting with finetuning specific multiagent system setups: \citet{subramaniam2025multiagent} propose \emph{Multiagent Finetuning}, establishing that training multiple agents outperforms training a single agent by preserving diverse reasoning chains; however, their approach uses supervised fine-tuning rather than RL, and focuses specifically on debate where homogeneous agents generate competing solutions to the same problem---a narrower setting than our sequential pipelines with heterogeneous specialized agents. \citet{liao2025marft} introduce \emph{MARFT}, providing theoretical foundations for adapting MARL to LLM-based multiagent systems; our work is complementary, focusing on the practical reward design challenge rather than algorithmic formalism. \citet{park2025maporl} present \emph{MAPoRL}, which co-trains agents through multi-turn debate where a verifier scores both answer correctness and discussion quality (rewarding corrective and persuasive exchanges). While MAPoRL demonstrates that co-training is essential---training individual LLMs alone fails to induce collaboration---their approach requires ground truth labels to train the verifier, uses sparse outcome-based rewards augmented with discussion incentives rather than dense per-action process feedback, and focuses on homogeneous debate agents rather than heterogeneous specialized pipelines. Our coach provides dense supervision at every action without requiring ground truth, enabling learning even when tasks fail entirely. \citet{liu2025magrpo} formalize LLM collaboration as a Dec-POMDP and propose \emph{MAGRPO}, extending GRPO to multiagent multi-turn settings; however, GRPO's same-state assumption breaks in heterogeneous pipelines where upstream stochasticity causes state divergence (see Section~\ref{sec:training}). We use REINFORCE++ with global batch normalization, and our coach's context-aware evaluation enables stable gradient estimation even when actions arise from divergent trajectories.

\subsection{Future Directions}

\textbf{Evaluation Limitations.} Our experiments use standard benchmark sizes: 32 AMC problems, 30 AIME 2025 problems, and 6 DSBench modeling tasks for held-out evaluation. While these sets are small, this reflects domain constraints---DSBench contains only 72 total modeling tasks, yielding an 8.6\% held-out ratio that is reasonable for such limited benchmarks. We prioritize training on AIME problems given their difficulty and relevance to mathematical reasoning. To mitigate variance from small evaluation sets, each problem is evaluated multiple times per checkpoint (4$\times$ for MathChat, 2$\times$ for DSBench) with accuracy computed as the mean across attempts. Due to computational constraints, we report results from single training runs rather than confidence intervals across multiple seeds. Tables report peak accuracy at the best-performing checkpoint for each benchmark, which may introduce optimistic bias. Future work should validate with multiple seeds as compute permits.

\textbf{Mitigating Coach Biases.} Using LLMs to assess LLM outputs has well-documented biases that affect evaluation reliability~\citep{ye2024justice, chen2024humans}. \emph{Verbosity bias} leads to preference for longer responses regardless of quality. \emph{Self-enhancement bias} emerges when LLMs favor outputs similar to their own generations~\citep{wataoka2024selfpreference}. Our results demonstrate the MAPPA is effective but such biases may still influence scores. In the future, we believe more investigation is warranted, with potential mitigation strategies including ensembling multiple coach models (e.g., combining Claude, Gemini, and GPT-4).

\textbf{Beyond Scalar Rewards.} Currently, our approach only uses scalar rewards from AI feedback, but the coaches are capable of much richer feedback. For instance, when an agent's action is suboptimal, the coach could generate a corrected action that the agents \textit{should have} taken, enabling supervised finetuning (SFT) on improved demonstrations or preference learning via DPO~\citep{rafailov2023dpo}, that combine these with RL. Extracting richer signal from each coach interaction could further improve sample efficiency, especially when agents perform a task to low success rate---even though this improvement may come at the expense of reduced exploration compared to pure RL as in our current work.

\textbf{Trainable Coach.}
\label{sec:trainable_coach}
The coach itself could be a trainable agent inside the multiagent system. This raises a fundamental question: what signal should train the coach? Options include meta-evaluation from a stronger external model, agreement with outcome-based verification when available, or human feedback on coach decisions. Whether a fully self-contained system---where agents and coaches co-evolve without external supervision---can avoid degenerate equilibria remains an open question.

\textbf{Reward Backpropagation.}
\label{sec:reward_backprop}
Our current approach assigns process rewards independently at each step---a bottom-up method where each agent is critiqued on everything that \emph{could} be improved, without knowing which improvements actually matter. A more efficient approach would \emph{backpropagate} outcome information: given a specific outcome, trace backward to identify which agent made which mistake, or which agent's action saved the team. At each backward step, the judge attempts to explain the outcome from the current agent's actions; if a reasonable explanation exists, credit or blame is attributed there, otherwise the judge peels off minor contributions and passes the residual to the previous agent. This mirrors formal backpropagation, where each layer provides a gradient direction for the layer below based on the current loss. We formalize this in Appendix~\ref{app:reward_backprop}.

\textbf{Agent-as-a-Coach.}
\label{sec:agent_coach}
Our current coach has full context about each agent's task, behavior, and consequences, but lacks awareness that its scores will be used to fine-tune models. This stateless evaluation underexploits LLM intelligence. A true \emph{agent-as-a-coach} would have access to training history and performance trends, enabling it to discover patterns like ``I have been scoring regression tasks +1.2 higher on average---am I creating imbalance?'' or ``classification success has stagnated while regression keeps improving.''

Beyond self-reflection, an agentic coach could implement \emph{strategic training}: ``First establish reliable task completion by rewarding successful runs even with mediocre quality; once success rates stabilize, shift focus to quality metrics.'' This mirrors how professional coaches build fundamentals before advanced techniques. The coach could use tools to compute statistics across training history, run code to verify correctness, or inspect intermediate artifacts---making evaluation decisions informed by the full training context rather than isolated action judgments. See Appendix~\ref{app:strategic_coach} for a detailed discussion of strategic multi-objective judging.

This paradigm shift from ``LLM-as-judge'' (passive, stateless) to ``agent-as-a-coach'' (active, strategic) scales with model capability: stronger coaches can implement more sophisticated curriculum strategies, detect and correct their own biases, and adaptively balance across task types based on observed training dynamics.

\textbf{Scaling to Large-Scale Multiagent Systems.} MAPPA simplifies end-to-end training of multiagent systems to crafting an effective LLM coach and providing it with relevant information to make accurate assessments. The true potential of this approach lies in scaling to large multiagent systems with dozens or more agents for complex, long-horizon tasks such as scientific research, where multiagent systems start to show promise~\citep{li2025freephdlabor}. A key challenge, however, is reward hacking: coach-assigned scores may improve while overall system success stagnates or declines. We believe collecting behavioral metrics throughout training, such as response length and tool call rate shown in Results, is key. These metrics serve as sanity checks and provide interpretable signals for diagnosing what is happening inside a multiagent system as it evolves.

%==============================================================================
\section{Conclusion}
%==============================================================================

MAPPA demonstrates that multiagent systems can be effectively trained end-to-end using process rewards from AI feedback. Dense per-action supervision solves credit assignment, improves sample efficiency, and generalizes across domains---suggesting that scaling specialized agents through finetuning, rather than prompting alone, represents a promising frontier for complex, long-horizon tasks. We are entering an era where autonomous AI agents interacting with each other is increasingly the default, from enterprise workflows to scientific discovery, with minimal human intervention. As these multiagent systems grow in capability and deployment, evaluating and improving them becomes a critical challenge. Our work takes a first step toward addressing this challenge.

%==============================================================================
\section*{Impact Statement}
%==============================================================================

This work advances methods for training multiagent AI systems with minimal human supervision, contributing to the broader goal of building self-improving AI systems. While this could accelerate progress on complex tasks that benefit society, training AI with AI supervisors raises value alignment concerns, as trained agents may inherit biases from coach models. We recommend auditing coach models for biases and monitoring behavioral metrics to detect reward hacking.

%==============================================================================
% References
%==============================================================================
\bibliography{references}
\bibliographystyle{icml2026}

%==============================================================================
% APPENDIX
%==============================================================================
\newpage
\appendix
\onecolumn

\section{Extended Experimental Results}
\label{app:extended_results}

\subsection{MathChat: Partial Information Results}
\label{app:mathchat_partial}

We also evaluate MAPPA under partial information constraints, where each agent observes only the immediately preceding agent's output (except the Problem Solver, which sees the problem statement), with no access to earlier context or the original question. This tests whether MAPPA can train agents that must operate with limited context.

\begin{table}[h]
\centering
\caption{MathChat performance under partial information constraints (Qwen3-4B). Despite degraded baseline performance compared to full context ($-$5--12pp), MAPPA achieves consistent improvements.}
\label{tab:mathchat_partial}
\small
\begin{tabular}{@{}lccc@{}}
\toprule
Benchmark & Baseline & Best & $\Delta$ \\
\midrule
AMC (32) & 72.7\% & 76.6\% & \up{+3.9pp} \\
AIME 2025 (30) & 37.5\% & 43.3\% & \up{+5.8pp} \\
\bottomrule
\end{tabular}
\end{table}

As expected, partial information constraints degrade baseline performance: AMC drops from 78.1\% to 72.7\% ($-$5.4pp) and AIME from 49.2\% to 37.5\% ($-$11.7pp) compared to full context. However, MAPPA still achieves consistent improvements of +3.9pp on AMC and +5.8pp on AIME, demonstrating that per-action process rewards provide effective supervision even when agents have limited observability.

\begin{figure*}[h]
\centering
\includegraphics[width=0.95\textwidth]{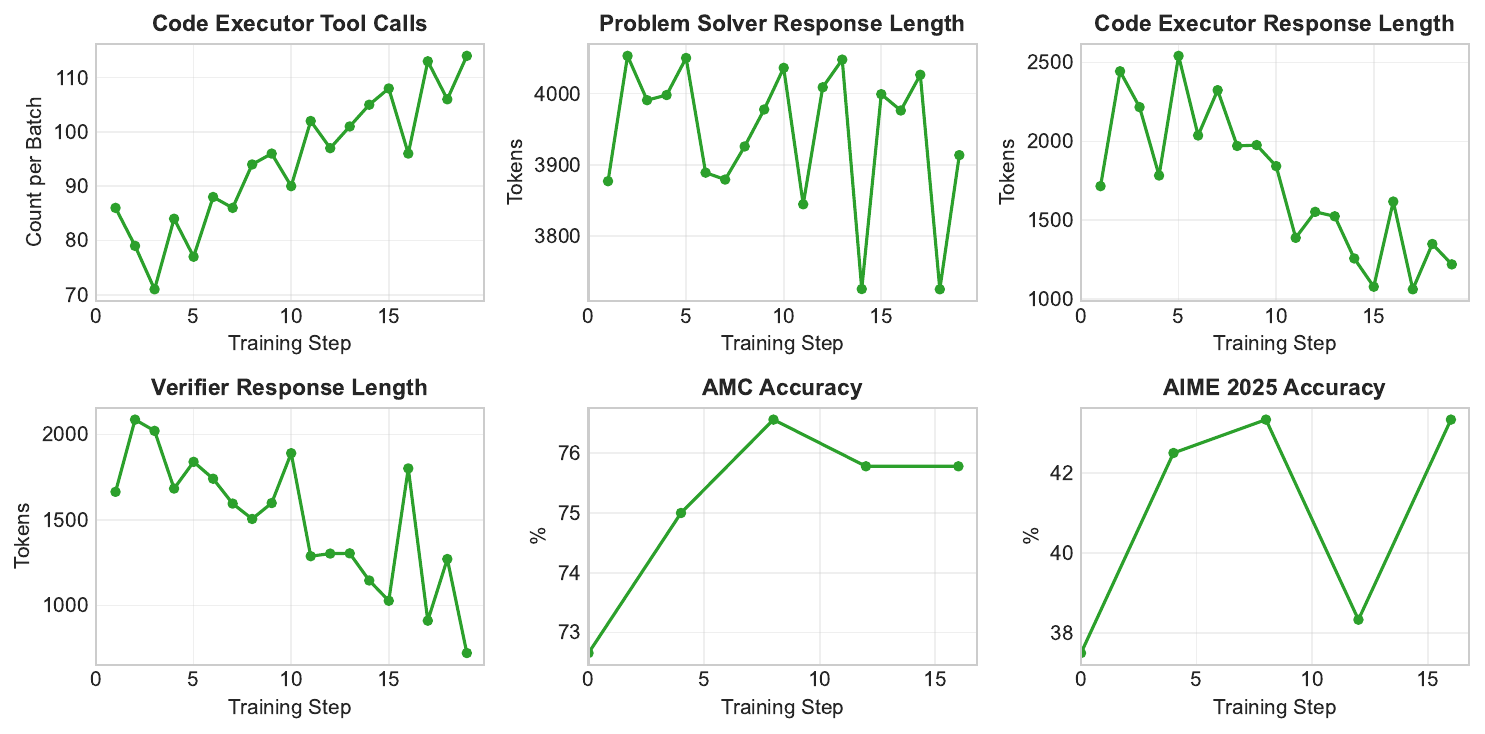}
\caption{Behavioral metrics for the partial information run (Qwen3-4B). From left to right, top to bottom: Code Executor tool calls, Problem Solver response length, Code Executor response length, Verifier response length, AMC accuracy, and AIME 2025 accuracy. Despite limited context, agents show similar behavioral adaptations as full-context runs.}
\label{fig:mathchat_behavioral_partial}
\end{figure*}

Figure~\ref{fig:mathchat_behavioral_partial} shows behavioral metrics for this run. Under partial information constraints, agents start with higher tool usage and shorter responses compared to full-context runs, then further optimize during training: tool calls increase while response lengths continue to decrease.

\subsection{DSBench: Extended Quality Metrics}
\label{app:dsbench_extended}

\begin{figure*}[h]
\centering
\includegraphics[width=0.95\textwidth]{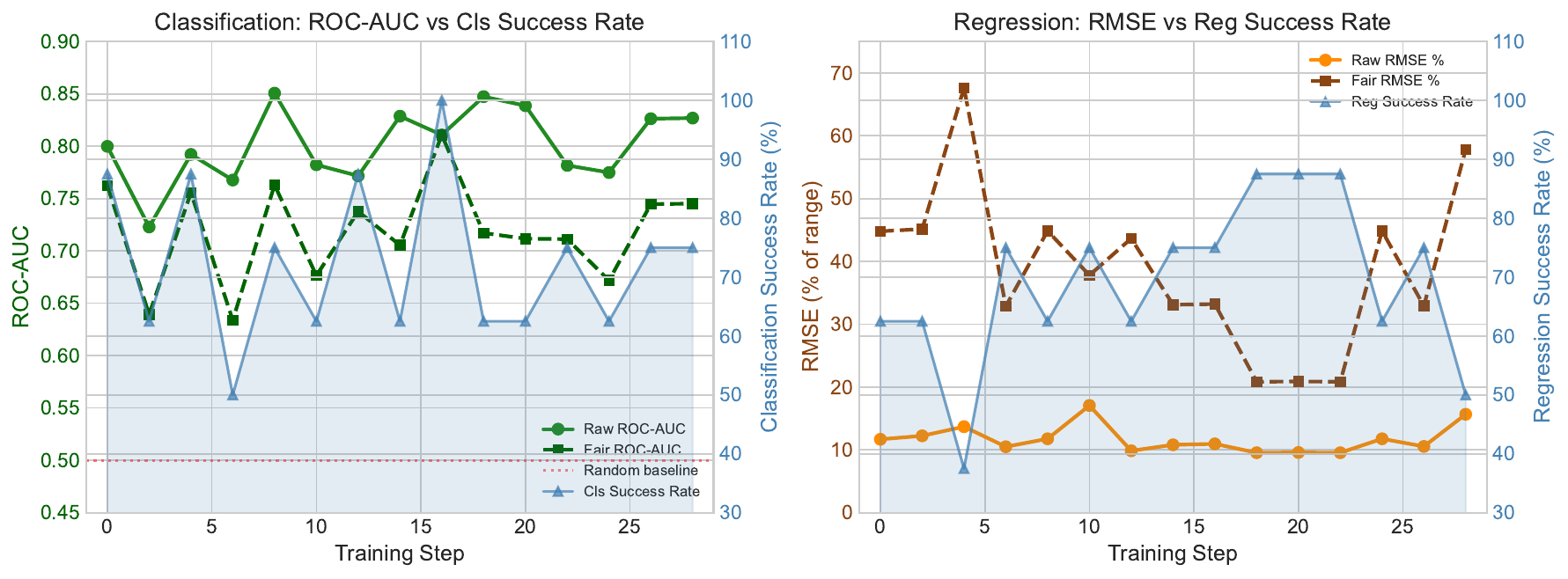}
\caption{DSBench quality metrics over 84 training steps. Top row: success rates by task type. Middle rows: classification metrics (Accuracy, F1, ROC-AUC) in raw and fair variants. Bottom row: regression metrics (MAE, RMSE) in raw and fair variants. Green vertical line and markers indicate step 44 (peak success rate). Fair metrics penalize failures (Accuracy: 0.5, F1: 0, MAE/RMSE: 50\%), capturing both quality and reliability.}
\label{fig:dsbench_quality_metrics}
\end{figure*}

Figure~\ref{fig:dsbench_quality_metrics} presents the complete set of quality metrics tracked during DSBench training, showing both raw metrics (computed only over successful samples) and fair metrics (which penalize failures: 0.5 for Accuracy, 0 for F1, 50\% for MAE/RMSE). The vertical green line marks step 44 (epoch 11), where overall success rate peaks.

Key observations: (1) Classification metrics (Accuracy, F1, ROC-AUC) peak around step 44 then decline, while regression metrics (MAE, RMSE) continue improving through step 84. (2) Fair metrics show larger improvements than raw metrics because they capture both quality improvement and increased success rate. (3) The divergence between raw and fair metrics after step 44 reflects the model's specialization toward regression tasks---raw classification metrics remain stable (successful runs maintain quality) while fair metrics decline (fewer successful runs).

\clearpage

%==============================================================================
\section{MathChat Agent Prompts}
\label{app:mathchat_prompts}
%==============================================================================

This section provides the complete prompts used by each agent in the MathChat pipeline. All agents receive the original math problem and role-specific instructions.

\subsection{Problem Solver Prompt}

\begin{small}
\begin{verbatim}
You are Problem Solver in a 3-agent system:
Problem Solver (you) -> Code Executor -> Verifier.

The system succeeds only if the Verifier (final agent)
outputs the correct answer. Your job is to draft a
solution to the problem.

You have a strict 4k token limit (your thinking inside
<think> and </think> tags also counts). Anything beyond
that will be truncated.

## Problem
{problem}
\end{verbatim}
\end{small}

\subsection{Code Executor Prompt}

\begin{small}
\begin{verbatim}
You are Code Executor in a 3-agent system:
Problem Solver -> Code Executor (you) -> Verifier.

The system succeeds only if the Verifier (final agent)
outputs the correct answer. Your job is to compute/verify
the solution using Python code.

You can execute Python code. Write code in ```python```
blocks and it will be automatically executed by the user
on your behalf, based on which you can iterate further
or output final answers.

You have a strict 4k token limit (your thinking inside
<think> and </think> tags also counts). Anything beyond
that will be truncated.

## Problem
{problem}

## Input from Problem Solver
{solution}
\end{verbatim}
\end{small}

\subsection{Verifier Prompt}

\begin{small}
\begin{verbatim}
You are Verifier, the final agent in a 3-agent system:
Problem Solver -> Code Executor -> Verifier (you).

You are the last agent. The system succeeds only if YOU
output the correct answer. Evaluate the information below
and provide the final answer.

You have a strict 4k token limit (your thinking inside
<think> and </think> tags also counts). Anything beyond
that will be truncated. Output your final answer as:
**\boxed{answer}**

## Problem
{problem}

## Input from Code Executor
{execution}
\end{verbatim}
\end{small}

\subsection{Coach Evaluation Prompt}

The coach evaluates each agent action using the following prompt template:

\begin{small}
\begin{verbatim}
You are evaluating a specific agent in a multiagent
system that collaboratively solves a math competition
problem. The system has 3 specialized agents working
sequentially:

1. **Problem Solver**: Reasons through the problem
   step-by-step
2. **Code Executor**: Writes and executes Python code
   to verify/compute solutions
3. **Verifier**: Synthesizes the code executor's output
   and provides the final answer.

The success of the overall system is determined by the
Verifier's final answer.

**Workflow Constraints:**
- Each agent has a 4K token output limit. Outputs beyond
  this are truncated.
- All agents can see the original problem statement.

When evaluating, consider whether the agent:
- (IMPORTANT) Performed well given their role and the
  information provided
- Stayed within output length limit

**Problem that the overall system is solving:**
{problem}

**Current agent being evaluated:**
{agent_role}

**What the agent received as input:**
{agent_input}

**What the agent outputted:**
{agent_output}

**Environment feedback for the agent's tool call
(if applicable):**
{tool_observation}

**Ground truth answer (if applicable):**
{ground_truth_answer}

Evaluate the agent's decision quality based on:
- Given the agent's role and the information it had as
  input, how helpful is its output towards the overall
  system's success?
- For Code Executor: Did it output code in the right
  format? Is the code syntactically correct and does it
  solve the problem?
- For Verifier: Did it correctly synthesize the
  information and output the final answer?

Strictly assigns a score from 0 (terrible) to 10 (perfect).

Additionally, if ground truth is provided (not "N/A") and
the agent is "Verifier", you MUST also verify answer
correctness:

Provide binary correctness as:
- ANSWER_CORRECT: 1 (if the Verifier's final answer
  matches the ground truth)
- ANSWER_CORRECT: 0 (if the Verifier's final answer
  is incorrect)

Structure your output in the following format exactly:

PROCESS_SCORE: [0 to 10]
ANSWER_CORRECT: [0 or 1, but only if ground truth
provided and agent is Verifier]
\end{verbatim}
\end{small}

\clearpage

%==============================================================================
\section{DSBench Agent Prompts}
\label{app:dsbench_prompts}
%==============================================================================

This section provides the complete prompts used by each agent in the DSBench pipeline. Prompts include placeholders for task-specific information and file system state.

\subsection{Data Engineer Prompt}

\begin{small}
\begin{verbatim}
You are a Data Engineer in a data science pipeline.

**CRITICAL: Keep your <think> reasoning BRIEF (under
1000 words). Then IMMEDIATELY write executable code in
a ```python block. Do NOT describe code in thinking -
just write it!**

**Task Description:**
{task_description}

## CRITICAL FILE SYSTEM OBSERVATION
**Files currently available in your workspace:**
{available_files_list}

NOTE: The file names shown above are the ACTUAL files
you can access. Always check this list!

**Your Responsibilities:**
1. **Load Data**: Read training data from 'train.csv'
   and test data from 'test.csv'
2. **Exploratory Data Analysis (EDA)**:
   - Inspect schema (dtypes, shape, missing values)
   - Analyze distributions, correlations
   - Identify data quality issues
3. **Data Preprocessing**:
   - Handle missing values (imputation/removal)
   - Encode categorical variables
   - Scale numerical features
   - Engineer new features based on domain insights
4. **CRITICAL: Save ALL Required Artifacts**:
   - `X_train.pkl` - preprocessed training features
   - `y_train.pkl` - training labels
   - `X_test.pkl` - preprocessed TEST features (CRITICAL!)
   - `scaler.pkl` or similar - fitted transformers
5. **Output Summary**: Print what files you saved

## CRITICAL: You MUST save X_test.pkl!
The Analyst agent downstream will load X_test.pkl to
generate predictions. If you don't save X_test.pkl,
the entire pipeline will FAIL.

**Constraints:**
- Apply SAME transformations to both train and test data
- Do NOT use test data for fitting (no data leakage!)
- Make preprocessing deterministic (set random_state)
- ALWAYS save X_test.pkl - downstream Analyst needs it!
\end{verbatim}
\end{small}

\subsection{Modeler Prompt}

\begin{small}
\begin{verbatim}
You are a Modeler in a data science pipeline.

**CRITICAL: Keep your <think> reasoning BRIEF (under
1000 words). Then IMMEDIATELY write executable code in
a ```python block.**

**Task Description:**
{task_description}

## CRITICAL FILE SYSTEM OBSERVATION
**Files currently available in your workspace (saved by
previous agents):**
{available_files_list}

NOTE: The file names shown above are the ACTUAL files
you can load. You MUST use these exact file names!

**Preprocessing Summary from Data Engineer:**
{preprocessing_summary}

**Your Responsibilities:**
1. **Check Available Files FIRST**: Verify which files
   exist before loading
2. **Load Data**: Load the preprocessed data saved by
   Data Engineer
3. **Algorithm Selection**:
   - Analyze the problem type (classification/regression)
   - Consider dataset characteristics
   - Select appropriate algorithms (RandomForest,
     XGBoost, LightGBM)
4. **Model Training**:
   - Train selected models with cross-validation
   - Tune hyperparameters
   - Evaluate performance
5. **Model Selection**: Choose best model based on
   validation metrics
6. **CRITICAL: Save Model**: Save the trained model as
   `model.pkl` - the Analyst needs this!
7. **Output**: Print files saved and performance metrics

**Constraints:**
- Check available files BEFORE trying to load them
- Set random_state for reproducibility
- Use cross-validation for model selection
- Report validation metrics honestly
- ALWAYS save your trained model to model.pkl
\end{verbatim}
\end{small}

\subsection{Analyst Prompt}

\begin{small}
\begin{verbatim}
You are an Analyst in a data science pipeline.

**CRITICAL: Keep your <think> reasoning BRIEF (under
1000 words). Then IMMEDIATELY write executable code.**

**Task Description:**
{task_description}

## CRITICAL FILE SYSTEM OBSERVATION
**Files currently available in your workspace:**
{available_files_list}

NOTE: The file names shown above are the ACTUAL files
you can load. You MUST use these exact file names!

**Preprocessing Summary from Data Engineer:**
{preprocessing_summary}

**Model Summary from Modeler:**
{model_summary}

**Your Responsibilities:**
1. **CHECK AVAILABLE FILES FIRST**: Look at the file
   list above CAREFULLY!
   - You NEED `model.pkl` to make predictions
   - You NEED `X_test.pkl` with preprocessed test features
   - If these files are MISSING, report the issue!
2. **Load Saved Artifacts**: Load the model and
   preprocessed test data
3. **Generate Predictions**: Make predictions on test set
4. **Get IDs from test.csv**: Load original test.csv
   to get the ID column
5. **Create Submission**:
   - Save predictions to 'submission.csv'
   - MUST include correct ID column and prediction column
   - Follow competition format
6. **Quality Checks**:
   - Validate no missing predictions
   - Check value ranges
   - Verify column names and order

## CRITICAL: Check file availability before loading!
If `X_test.pkl` or `model.pkl` is NOT in the available
files list, you CANNOT create submission.csv.

**Constraints:**
- CHECK THE FILE LIST ABOVE before trying to load
- If required files are missing, report which ones
- Load the model from model.pkl
- Load preprocessed test data from X_test.pkl
- Get IDs from original test.csv
- Ensure predictions match test set size
- Save predictions to submission.csv
\end{verbatim}
\end{small}

\subsection{Coach Evaluation Prompt}

The DSBench coach uses a detailed prompt that includes root-cause analysis for pipeline failures. Key excerpts:

\begin{small}
\begin{verbatim}
You are evaluating a specific agent in a multiagent
data science system. The system has 3 specialized agents:

1. **Data Engineer**: Performs EDA, data cleaning,
   feature engineering
   - MUST save: X_train.pkl, y_train.pkl, X_test.pkl
2. **Modeler**: Selects algorithms, trains models
   - MUST save: model.pkl
3. **Analyst**: Generates predictions, creates submission
   - MUST save: submission.csv (FINAL DELIVERABLE)

## CRITICAL: OUTCOME-BASED ROOT CAUSE ANALYSIS

If ground_truth_answer shows "ERROR: No predictions
found" or similar failure, this means submission.csv
was NOT created. This is a PIPELINE FAILURE.

You MUST perform root cause analysis:

1. **Check "FILES SAVED BY PREVIOUS AGENTS"**:
   - Did Data Engineer save X_test.pkl?
   - Did Modeler save model.pkl?
   - If these files are missing, the UPSTREAM AGENT
     is at fault!

2. **Check "tool_observation" for errors**:
   - FileNotFoundError for X_test.pkl -> Data Engineer's
     fault
   - FileNotFoundError for model.pkl -> Modeler's fault
   - KeyError, TypeError -> Likely Analyst's own bug

3. **Assign blame to the responsible agent**:
   - Data Engineer's fault if: X_test.pkl is NOT in
     "FILES SAVED BY PREVIOUS AGENTS"
   - Modeler's fault if: model.pkl is NOT in
     "FILES SAVED BY PREVIOUS AGENTS"
   - Analyst's fault if: Both files were available but
     Analyst's code is buggy

## Scoring Guidelines

**Score 10**: EXCEPTIONAL - Agent exceeded expectations
**Score 8-9**: Code executed successfully, all required
  files saved, methodology is sound
**Score 6-7**: Code executed, most files saved, minor
  issues
**Score 4-5**: Agent tried but code had errors
**Score 2-3**: Code crashed due to upstream failures
**Score 0-1**: Agent produced no useful output

## Using Ground Truth Metrics (Analyst Only)

If ground truth metrics are provided, apply HARD CAPS:

**Classification Tasks:**
| ROC-AUC >= 0.85 | Max 9 |
| ROC-AUC 0.75-0.85 | Max 8 |
| ROC-AUC 0.65-0.75 | Max 6 |
| ROC-AUC 0.55-0.65 | Max 4 |
| ROC-AUC < 0.55 | Max 3 |

**Regression Tasks (RMSE as % of range):**
| RMSE < 10% | Max 9 |
| RMSE 10-25% | Max 8 |
| RMSE 25-50% | Max 6 |
| RMSE > 50% | Max 4 |

Output format:
PROCESS_SCORE: [0 to 10]
\end{verbatim}
\end{small}

\clearpage

%==============================================================================
\section{Example Coach Evaluations}
\label{app:coach_examples}
%==============================================================================

This section presents real coach evaluation responses from training runs, illustrating how the coach provides dense, context-aware feedback that goes beyond simple correctness checking.

\subsection{MathChat: Problem Solver Evaluation}

The following coach response evaluates a Problem Solver agent attempting an AMC geometry problem about counting rectangles inscribed in a regular dodecagon. This example demonstrates how the coach rewards metacognitive monitoring---the agent's ability to test and revise its own hypotheses.

\begin{small}
\begin{verbatim}
PROCESS_SCORE: 0.8

**Evaluation:**

1. **Reasonableness of the Action:** The agent's action,
   which is to provide a step-by-step reasoning process,
   is perfectly aligned with its role as Problem Solver.
   The reasoning itself is logical and demonstrates a
   good attempt to solve the problem.

2. **Quality of Reasoning:**
   * **Initial Insight:** The agent has a crucial insight
     early on: a rectangle inscribed in a circle (which
     the dodecagon is) must have its diagonals as
     diameters of the circle. This is the correct path
     to the solution.
   * **Flawed Premise:** The agent then makes a common
     mistake, assuming that for the quadrilateral to be
     a rectangle, the two diameters must be perpendicular.
     This is incorrect; the quadrilateral formed by the
     endpoints of *any* two distinct diameters is always
     a rectangle.
   * **Excellent Self-Correction:** The agent then
     intelligently uses the example provided in the asy
     code to test its own hypothesis. It correctly
     identifies the vertices of the example rectangle
     and determines that its diagonals are not
     perpendicular. This leads it to correctly conclude
     that its premise was wrong. This act of testing a
     hypothesis against evidence and revising its
     thinking is a sign of a very strong reasoning
     process.
   * **Getting Stuck:** After correctly abandoning the
     "perpendicular diameters" idea, the agent
     unfortunately also abandons the "diameters" idea
     altogether. It switches to a much more complex
     method of analyzing perpendicular sides and gets
     stuck. It failed to realize that its original
     insight was correct and only the "perpendicular"
     constraint needed to be removed.

**Conclusion:**

The agent demonstrated a strong, logical thought process.
It identified the most promising approach, made a mistake,
but then impressively corrected that mistake based on the
provided evidence. Its main failure was not revisiting
the original, correct approach after removing the flawed
constraint. Despite not reaching the final answer, the
quality of the step-by-step reasoning and self-correction
is high.
\end{verbatim}
\end{small}

This example illustrates how the coach rewards metacognitive monitoring: when the agent tests its hypothesis against evidence and revises its thinking, this is explicitly praised as ``a sign of a very strong reasoning process.'' The coach also distinguishes between the quality of the initial insight (correct), the intermediate error (flawed premise), the recovery attempt (excellent self-correction), and the final outcome (got stuck). This granular feedback enables agents to learn which reasoning behaviors to reinforce.

\subsection{DSBench: Data Engineer Evaluation}

The following coach response evaluates a Data Engineer agent preprocessing a healthcare dataset for stroke prediction. This example demonstrates how the coach identifies subtle implementation bugs that would silently corrupt the pipeline.

\begin{small}
\begin{verbatim}
PROCESS_SCORE: 0.6

**Evaluation of agent_data_engineer:**

**1. Reasoning and Plan:**
The agent's thought process is logical and structured. It
correctly identifies the key responsibilities: loading
data, performing EDA, handling missing data, encoding
categoricals, scaling numericals, and engineering new
features. The agent correctly reasons about the need for
deterministic preprocessing and considers the appropriate
order of operations.

**2. Code Implementation and Preprocessing:**

* **Strengths:**
  - The use of sklearn.pipeline.Pipeline and
    sklearn.compose.ColumnTransformer is a best practice.
    It encapsulates preprocessing steps cleanly and helps
    prevent data leakage.
  - The choice of preprocessing steps is appropriate:
    mean imputation for numerical, constant fill for
    categorical, StandardScaler for numerical, and
    OneHotEncoder for categorical.
  - The agent correctly includes handle_unknown='ignore'
    in the OneHotEncoder, which is crucial for handling
    potential new categories in the test set.
  - The agent correctly sets random_state=42 to ensure
    the process is deterministic.
  - The feature engineering ideas (age_group, bmi_over_age)
    are plausible and relevant for a stroke prediction
    task.

* **Weaknesses / Bugs:**
  - **Critical Flaw:** The agent defines numerical_cols
    and categorical_cols *before* creating the new
    features (age_group, bmi_over_age). Consequently, the
    ColumnTransformer is configured to only process the
    *original* columns. The newly created features will
    be dropped by the transformer, completely nullifying
    the feature engineering step. The column lists should
    have been defined *after* the new columns were added
    to the DataFrame.
  - **Redundancy:** The agent performs imputation twice.
    First, it imputes directly on the DataFrame using
    .fillna(). Then, it includes SimpleImputer again
    within the pipelines of the ColumnTransformer. While
    not a fatal error, it is inefficient and indicates a
    slight misunderstanding of how to use the pipelines
    effectively.

**3. Summary Output:**
The agent correctly structures the final output according
to the specified PREPROCESSING_SUMMARY format.

**Conclusion:**
The agent demonstrates a strong conceptual understanding
of the data engineering role and utilizes modern, robust
tools (sklearn pipelines). However, the implementation
contains a critical logical error in the order of
operations that prevents the engineered features from
being correctly processed. This is a significant bug that
would negatively impact the final model. The plan was
excellent, but the execution was flawed.
\end{verbatim}
\end{small}

This example demonstrates the coach's ability to identify subtle implementation bugs that would silently corrupt the ML pipeline. The ``Critical Flaw''---defining column lists before creating new features, causing engineered features to be silently dropped---is exactly the kind of bug that passes syntax checks but produces incorrect results. The coach also evaluates domain-specific best practices: using \texttt{ColumnTransformer} for clean preprocessing, setting \texttt{handle\_unknown='ignore'} for robust handling of unseen categories, and ensuring deterministic behavior with \texttt{random\_state}. The conclusion ``the plan was excellent, but the execution was flawed'' provides actionable feedback that helps the agent learn to verify implementation details match high-level intentions.

\subsection{MathChat: Strategy vs. Implementation}

The following example evaluates a Problem Solver on a grid coloring combinatorics problem, demonstrating how the coach distinguishes between excellent high-level strategy and flawed low-level modeling.

\begin{small}
\begin{verbatim}
PROCESS_SCORE: 0.6

This is a very strong attempt at a complex combinatorial
problem. The agent's overall strategy is excellent, but
there is a critical flaw in its understanding of the
grid's structure that prevents it from reaching the
correct conclusion.

**Evaluation of the Agent's Reasoning:**

1. **Problem Decomposition (Excellent):** The agent
   correctly identifies the 12 line segments and astutely
   categorizes them into 8 "outer" segments (on the
   perimeter) and 4 "internal" segments (shared between
   squares). This is a key insight that simplifies the
   problem.

2. **Strategic Approach (Excellent):** The agent devises
   a powerful strategy:
   - First, consider all possible colorings of the 4
     internal segments.
   - Second, for each internal coloring, calculate the
     number of ways the 8 outer segments can be colored
     to satisfy the constraints for each of the four
     squares.
   - Finally, sum the results over all possible internal
     colorings.
   This is a systematic and valid way to solve the
   problem.

3. **Modeling the Grid Structure (Flawed):** This is
   where the agent makes a critical error. It incorrectly
   assumes the grid can be split into two independent
   columns.
   - **Agent's flawed model:** It states that the
     top-left and bottom-left squares share two internal
     edges, and the top-right and bottom-right squares
     share the other two internal edges. This would mean
     the left column is completely independent of the
     right column.
   - **Correct model:** The internal edges connect all
     four squares. For example, the top-middle vertical
     segment is shared by the top-left and top-right
     squares, linking the two columns.

4. **Execution of the Flawed Model (Good):** Given its
   incorrect assumption, the agent's subsequent reasoning
   is logical and well-executed. It correctly sets up
   the calculation based on its flawed model. If the
   initial model had been correct, this method would
   have yielded the right answer for that model.

**Conclusion:**

The agent demonstrated a sophisticated approach to
problem-solving. The strategy of breaking down the
problem based on the more constrained "internal" edges
is exactly the right way to tackle this. However, the
fundamental misunderstanding of the grid's connectivity
is a major flaw that makes the final answer incorrect.
The high-level strategy is excellent, but the low-level
implementation is based on a false premise.
\end{verbatim}
\end{small}

This example shows the coach's nuanced evaluation: ``Problem Decomposition (Excellent),'' ``Strategic Approach (Excellent),'' but ``Modeling the Grid Structure (Flawed).'' The coach explicitly notes that the agent's execution of its flawed model was internally consistent---``if the initial model had been correct, this method would have yielded the right answer.'' This distinction between strategic quality and modeling accuracy provides precise feedback for improvement.

\subsection{DSBench: Analyst Pipeline Failure}

The following example evaluates an Analyst agent that fundamentally misunderstands how to apply a trained ML pipeline to test data---a critical conceptual error in production ML systems.

\begin{small}
\begin{verbatim}
PROCESS_SCORE: 0.1

**Critique of the Agent's Action:**

1. **Reproducing the Pipeline (Major Failure)**: The
   agent's primary responsibility is to apply the *exact*
   same preprocessing and use the *exact* same trained
   model from the previous steps. The agent fails
   critically on this point:
   - **LabelEncoder Error**: The code le = LabelEncoder()
     and le.fit_transform(test_df['...']) creates a *new*
     encoder and fits it *only on the test data*. This is
     incorrect. The encoder must be the one that was
     fitted on the training data to ensure consistent
     mapping of categories to numbers.
   - **TfidfVectorizer Error**: The code initializes a
     new, unfitted vectorizer. Calling .transform()
     without first calling .fit() on the training data
     will cause a NotFittedError. The agent should have
     used the vectorizer object that was already fitted
     by the Data Engineer/Modeler.
   - **Model Retraining Error**: The code re-initializes
     a new RandomForestClassifier and then attempts to
     retrain it with model.fit(X_train, y_train). This
     is a severe misunderstanding of the agent's role.
     The Analyst should use the *already trained* model
     from the Modeler. Furthermore, the variables
     X_train and y_train are not defined within the
     Analyst's script, meaning the code would crash
     with a NameError.

2. **Generating Predictions (Failure)**: Due to the
   errors mentioned above, the provided script is not
   executable. It would fail at the tfidf.transform
   step, and even if it didn't, it would fail at the
   model.fit step. Therefore, it cannot generate any
   predictions.

3. **Creating Submission (Partial Success)**: The logic
   for creating the final submission DataFrame and
   printing it in the required format is syntactically
   correct. However, this is useless since the
   predictions themselves cannot be generated.

**Conclusion:**

The agent demonstrated a fundamental misunderstanding of
how to apply a trained machine learning pipeline to new,
unseen data. The concepts of using pre-fitted transformers
(like LabelEncoder and TfidfVectorizer) and a pre-trained
model were completely missed. The resulting code is
non-functional and conceptually wrong.
\end{verbatim}
\end{small}

This example demonstrates the coach identifying fundamental conceptual errors about ML deployment. The agent's mistake---refitting encoders on test data and retraining the model---would cause category mapping inconsistencies and invalidate the entire pipeline. This is precisely the kind of error that distinguishes production-ready ML practitioners from novices. The coach's detailed breakdown (LabelEncoder error, TfidfVectorizer error, Model Retraining error) provides specific, actionable feedback about each conceptual gap.

\clearpage

%==============================================================================
\section{Training Algorithm Details}
\label{app:training}
%==============================================================================

This appendix provides additional technical details for the training algorithm described in Section~\ref{sec:training}.

\subsection{Why REINFORCE++ Over GRPO}

Group-Relative Policy Optimization (GRPO) normalizes advantages within groups of samples sharing the same prompt, which assumes identical input states across the group. This assumption is valid for single-agent settings where all samples from the same prompt begin in the same state. However, in end-to-end multiagent training, each agent's input context depends on the stochastic output of upstream agents. Consider two rollouts of the same math problem: the Problem Solver may generate different reasoning chains, causing the Code Executor to observe different inputs despite originating from the same initial prompt. This state divergence makes within-group comparisons ill-defined for GRPO.

An alternative approach---training each agent separately with frozen upstream context---would restore the same-state assumption. However, this requires separate rollout passes for each agent, significantly reducing efficiency and preventing agents from co-adapting during training. Instead, we train all agents end-to-end from complete trajectories using REINFORCE++, which applies global batch normalization across all collected experiences rather than per-prompt normalization.

\subsection{Advantage Estimation}

The advantage estimation proceeds in two steps. First, for each action $a_t$ with coach reward $r_t^{\text{coach}} \in [0, 1]$, we compute the KL-penalized reward:
\begin{equation}
r_t = r_t^{\text{coach}} - \beta \cdot D_{\text{KL}}(\pi_\theta \| \pi_{\text{ref}})
\end{equation}
where $\beta = 0.01$ is the KL penalty coefficient and $D_{\text{KL}}$ measures divergence from the frozen reference policy. The advantage for each action is then computed as undiscounted return-to-go:
\begin{equation}
A_t = \sum_{\tau \geq t} r_\tau
\end{equation}
which propagates downstream rewards back to earlier actions within each agent's trajectory. This choice of $\gamma = 1$ (no discounting) reflects the finite-horizon nature of multiagent workflows where all actions contribute equally to the final outcome.

Second, we collect advantages $A_t$ from all agents and all experiences in the current batch, then compute global statistics:
\begin{equation}
\mu = \frac{\sum_{i,t} A_t^{(i)} \cdot m_t^{(i)}}{\sum_{i,t} m_t^{(i)}}, \quad
\sigma^2 = \frac{\sum_{i,t} (A_t^{(i)} - \mu)^2 \cdot m_t^{(i)}}{\sum_{i,t} m_t^{(i)}}
\end{equation}
where $m_t^{(i)}$ are action masks indicating valid actions. Advantages are then normalized: $\hat{A}_t = (A_t - \mu) / \sqrt{\sigma^2 + \epsilon}$ with $\epsilon=10^{-8}$ for numerical stability.

\subsection{Policy Gradient Objective}

The policy gradient objective follows PPO with clipped surrogate:
\begin{equation}
\mathcal{L}_{\text{policy}} = -\mathbb{E}\! \left[ \min\!\left( \rho_t(\theta) \hat{A}_t, \text{clip}(\rho_t(\theta), 1\!-\!\epsilon, 1\!+\!\epsilon) \hat{A}_t \right) \right]
\end{equation}
where $\rho_t(\theta) = \pi_\theta(a_t|s_t)/\pi_{\text{old}}(a_t|s_t)$ is the probability ratio between the current and old policy, and $\epsilon = 0.2$ is the clipping range. The clipping prevents excessively large policy updates that could destabilize training, which is particularly important when training multiple agents simultaneously where instability in one agent can cascade through the pipeline.

\subsection{Strategic Multi-Objective Judging}
\label{app:strategic_coach}

Our current coach evaluates each action independently without memory of previous training epochs or access to aggregate performance statistics. Each action is scored based solely on its intrinsic quality, without considering whether the training process would benefit more from improving reliability versus refinement at that stage.

A strategic coach with memory could balance objectives dynamically. For example, such a coach could:
\begin{itemize}[nosep]
    \item Assign higher rewards to successful task completion early in training when the pipeline is unreliable
    \item Shift focus to quality improvements once success rate stabilizes
    \item Restore emphasis on reliability if it degrades
\end{itemize}

Such a coach could access metrics like rolling success rate, quality trends, and per-agent failure modes to make context-aware reward decisions across the training trajectory. Implementing strategic coaching in our framework would be straightforward: the coach prompt already accepts arbitrary metadata fields that could include per-epoch statistics. We leave implementation and empirical evaluation of strategic multi-objective judging to future work.

\clearpage

%==============================================================================
\section{Reward Backpropagation}
\label{app:reward_backprop}
%==============================================================================

This appendix formalizes \emph{Reward Backpropagation}, an approach for outcome-aware process reward decomposition referenced in Section~\ref{sec:reward_backprop}.

\subsection{Motivation}

Process rewards without outcome information are inherently inefficient: without knowing what actually mattered, a coach must critique everything that \emph{could} be improved---some feedback may be valid, some may be mere stylistic preference. This bottom-up approach asks each agent to ``improve my part to my best effort'' without knowing which improvements actually matter for the final result.

Outcome-aware evaluation inverts this. Given a specific outcome (e.g., incorrect predictions, or surprisingly good results despite upstream errors), it becomes much easier to trace backward and identify which agent made which mistake, or whose action saved the team. This top-down approach asks: ``given this outcome, who is the real contributor to success or failure?'' The linear backpropagation structure provides an ordered flow of attribution---not a chaotic discussion---where each step receives credit or blame only for the residual not yet explained by downstream actions.

Reward Backpropagation formalizes this shift from local, bottom-up optimization to global-aware, top-down attribution.

\subsection{Method Overview}

Given a trajectory $\tau = (s_1, s_2, \ldots, s_T)$ and final outcome $R_T \in \mathbb{R}^k$ (potentially vector-valued), Reward Backpropagation proceeds in two phases:

\paragraph{Forward Pass (Process Rewards).} For each step $t$, query a coach to produce a process reward $p_t$ evaluating local decision quality without knowledge of the final outcome.

\paragraph{Backward Pass (Outcome Decomposition).} Starting from $R_T$, run a backward chain of coach calls:
\begin{enumerate}[nosep]
    \item At step $T$, the coach attributes part of $R_T$ to step $T$, producing contribution $\Delta_T$ and passing residual $r_{T-1}$ backward
    \item At step $t < T$, the coach receives residual $r_t$, attributes part to step $t$ (producing $\Delta_t$), and passes $r_{t-1} = r_t - a_t$ backward
    \item This continues to step 1, yielding per-step outcome-aware contributions $\{\Delta_t\}$
\end{enumerate}

\textbf{Implementation Note.} The notation above is \emph{conceptual} rather than literal. In practice, the ``residual'' $r_t$ and ``subtraction'' $r_{t-1} = r_t - a_t$ are not scalar or vector arithmetic but rather \emph{natural language attribution}. Each backward step asks the coach to explain, in natural language, how the output of agent $t-1$ (which forms part of the input to agent $t$) contributed to the outcome being evaluated. For example, the coach might produce feedback such as: ``Agent $t-1$'s output contained attribute X, which caused issue Y at agent $t$. I consider X to be a problematic decision, worth $-3$ out of 10.'' The ``subtraction'' thus represents conceptual decomposition---the coach articulates what portion of the outcome can be attributed to each step, with the residual being the \emph{remaining unexplained outcome} passed backward for further attribution. This natural language grounding makes the process interpretable and allows rich, multi-dimensional feedback beyond what scalar rewards can capture.

\subsection{Combining Process and Outcome Signals}

The final step-level reward combines both signals:
\begin{equation}
f_t = \alpha \cdot p_t + \beta \cdot \Delta_t
\end{equation}
where $\alpha, \beta \geq 0$ are hyperparameters. Process reward $p_t$ captures local quality; outcome contribution $\Delta_t$ captures how the final result validates or refutes that quality. Steps whose local quality did not affect the outcome receive small $|\Delta_t|$.

\subsection{Consistency Constraints}

We encourage approximate decomposition consistency:
\begin{equation}
\sum_{t=1}^T \Delta_t \approx v(R_T)
\end{equation}
where $v: \mathbb{R}^k \to \mathbb{R}$ scalarizes the outcome vector. This can be enforced via prompting or post-hoc normalization.

\subsection{Benefits}

Reward Backpropagation provides: (1) causally plausible credit assignment---locally good actions in failed trajectories receive negative $\Delta_t$; (2) interpretable narratives about which steps mattered; (3) robustness to process-reward hacking since $\Delta_t$ anchors rewards to actual outcomes. The approach is compatible with REINFORCE++ and can be distilled into a lightweight Process Reward Model for efficiency.

\clearpage

%==============================================================================
\section{Distributed Training Implementation}
\label{app:implementation}
%==============================================================================

This appendix provides detailed implementation specifications for the distributed training architecture described in Section~\ref{sec:implementation}.

\subsection{Parallel Agent Initialization}

Each agent is initialized via \texttt{ThreadPoolExecutor} to parallelize model loading across agents:

\begin{small}
\begin{verbatim}
with ThreadPoolExecutor(max_workers=len(agent_configs)) as executor:
    futures = [executor.submit(self._init_agent, agent_id, config, args)
               for agent_id, config in agent_configs]
    agents = [f.result() for f in futures]
\end{verbatim}
\end{small}

Each agent instantiates independent Ray actor groups for:

\begin{itemize}[nosep]
    \item \textbf{vLLM Engines}: \texttt{num\_engines} $\times$ \texttt{tensor\_parallel\_size} Ray actors for autoregressive generation with PagedAttention and KV-cache optimization
    \item \textbf{Actor Policy}: \texttt{PPORayActorGroup} with \texttt{actor\_num\_nodes} $\times$ \texttt{actor\_num\_gpus\_per\_node} workers for trainable policy parameters
    \item \textbf{Reference Model}: Same structure as actor, frozen copy for KL divergence computation
    \item \textbf{Critic Model}: Value function estimation distributed across ranks (optional, not used with REINFORCE++)
\end{itemize}

When \texttt{colocate\_all\_models=True}, placement groups bundle multiple models on shared GPUs with fractional allocation (\texttt{num\_gpus\_per\_actor=0.2}), reducing peak memory by enabling actor, reference, and vLLM models to share GPU memory.

\subsection{Distributed Rollout Generation}

Prompts are sharded evenly across \texttt{world\_size} workers:

\begin{small}
\begin{verbatim}
chunk_size = (num_prompts + world_size - 1) // world_size
for rank in range(world_size):
    start_idx, end_idx = rank * chunk_size, min((rank+1) * chunk_size, num_prompts)
    chunked[rank] = prompts[start_idx:end_idx]
\end{verbatim}
\end{small}

Each rank receives its shard and executes workflows via Ray remote calls:

\begin{small}
\begin{verbatim}
all_refs = []
for rank in range(world_size):
    ref = generate_samples_remote.remote(world, chunked[rank], rank, world_size)
    all_refs.append(ref)
results = ray.get(all_refs)  # Barrier: wait for all ranks
\end{verbatim}
\end{small}

Within each workflow, LLM engines are assigned cyclically: if fewer engines than ranks, engine $i$ handles rank $i \mod \text{num\_engines}$; otherwise engines are striped across ranks. Requests are batched across available engines:

\begin{small}
\begin{verbatim}
batch_size = (len(prompts) + len(llms) - 1) // len(llms)
for i, llm in enumerate(llms):
    batch = prompts[i * batch_size : (i + 1) * batch_size]
    refs.append(llm.generate.remote(batch, sampling_params))
\end{verbatim}
\end{small}

\subsection{Experience Preparation and Agent Routing}

Completed trajectories are processed by \texttt{ExperienceMaker}, which computes per-action rewards, KL divergence, and advantages. For multiagent training, each action is routed to its originating agent:

\begin{small}
\begin{verbatim}
sharded_data = [[None] * world_size for _ in range(num_agents)]
for rank in range(world_size):
    for agent_id in range(num_agents):
        agent_data = all_results[rank][agent_id]  # Extract agent's batch
        sharded_data[agent_id][rank] = ray.put(agent_data)
\end{verbatim}
\end{small}

This enables each agent to train on its own trajectory segments while maintaining batch synchronization across the distributed system.

\paragraph{Variable Turn Count Handling.}
When agents have variable turn counts (e.g., Code Executor in MathChat with up to 5 turns), different workers may produce different numbers of samples, causing NCCL deadlocks during gradient synchronization. The \texttt{filter\_agents\_data} option truncates all workers to the minimum sample count:

\begin{small}
\begin{verbatim}
# Count actual samples per rank for each agent
actual_samples_per_rank = [[count_samples(rank, agent)
                            for agent in range(num_agents)]
                           for rank in range(world_size)]

# Truncate to minimum across ranks to ensure equal gradient syncs
min_samples = [min(rank[agent] for rank in actual_samples_per_rank)
               for agent in range(num_agents)]
\end{verbatim}
\end{small}

This is critical for MathChat where \texttt{coder\_max\_turns=5} causes variable action counts across workers.

\paragraph{Global Batch Normalization (REINFORCE++).}
Advantages are normalized across the entire batch, not per-prompt:

\begin{small}
\begin{verbatim}
all_advantages = torch.cat([exp.advantages for exp in experiences])
mean = (all_advantages * action_masks).sum() / num_actions
var = ((all_advantages - mean).pow(2) * action_masks).sum() / num_actions
rstd = var.clamp(min=1e-8).rsqrt()
normalized = (all_advantages - mean) * rstd
\end{verbatim}
\end{small}

\subsection{Weight Synchronization}

After each training step, updated weights must be broadcast from DeepSpeed actors to vLLM engines. The synchronization mechanism depends on \texttt{colocate\_all\_models}:

\paragraph{Standard NCCL Broadcast.} When models are on separate GPUs, a dedicated NCCL process group connects DeepSpeed rank 0 with all vLLM engine ranks:

\begin{small}
\begin{verbatim}
# Process group: [DeepSpeed rank 0, vLLM engine 0, engine 1, ...]
world_size = vllm_num_engines * vllm_tensor_parallel_size + 1

# Broadcast each parameter (ZeRO-3 requires AllGather first)
for name, param in model.named_parameters():
    with deepspeed.zero.GatheredParameters([param], enabled=zero_stage==3):
        torch.distributed.broadcast(param.data, src=0, group=model_update_group)
\end{verbatim}
\end{small}

\paragraph{CUDA IPC (Co-located Models).} When \texttt{colocate\_all\_models=True} (default for both MathChat and DSBench), the system automatically selects CUDA IPC for zero-copy parameter sharing, which is faster than NCCL for models on the same GPU:

\begin{small}
\begin{verbatim}
# Automatically enabled when colocate_all_models=True and backend="nccl"
weight = param.data.clone()
ipc_handle = reduce_tensor(weight)  # Get CUDA IPC handle
ipc_handle = {get_physical_gpu_id(): ipc_handle}
torch.distributed.all_gather_object(ipc_handle_list, ipc_handle)
# vLLM engines reconstruct tensor from IPC handle without data copy
\end{verbatim}
\end{small}

\subsection{Memory Optimization}

\paragraph{Prefix Caching.} vLLM reuses KV cache across requests with common prefixes (e.g., system prompts shared across agents). Cache is reset after weight updates: \texttt{engine.reset\_prefix\_cache.remote()}.

\paragraph{Sleep/Wake Mode.} vLLM engines can offload to CPU during training to free GPU memory:

\begin{small}
\begin{verbatim}
def sleep(self, level=1):
    self.llm.sleep(level=level)  # Offload to CPU
def wake_up(self):
    self.llm.wake_up()           # Reload to GPU
\end{verbatim}
\end{small}

\paragraph{DeepSpeed ZeRO-3.} Optimizer states, gradients, and parameters are partitioned across GPUs. Parameters are gathered on-demand during forward/backward passes, enabling training of models larger than single-GPU memory.

\subsection{Synchronization Points}

Table~\ref{tab:sync_points} summarizes the key synchronization barriers in the training loop.

\begin{table}[h]
\centering
\caption{Communication patterns and synchronization points in distributed training.}
\label{tab:sync_points}
\small
\begin{tabular}{@{}lll@{}}
\toprule
Component & Protocol & Frequency \\
\midrule
Prompt sharding & Ray \texttt{put}/\texttt{get} & Per iteration \\
Rollout collection & Ray \texttt{remote}/\texttt{get} & Per iteration \\
Reference model inference & Ray \texttt{remote} & Per trajectory \\
Gradient reduction & NCCL \texttt{all\_reduce} & Per training step \\
Weight broadcast & NCCL \texttt{broadcast} & Per training step \\
Metric aggregation & NCCL \texttt{all\_reduce} & Per training step \\
\bottomrule
\end{tabular}
\end{table}

\subsection{Configuration Parameters}

Table~\ref{tab:config_params} lists key hyperparameters controlling distributed training behavior.

\begin{table}[h]
\centering
\caption{Key configuration parameters for distributed training.}
\label{tab:config_params}
\small
\begin{tabular}{@{}lll@{}}
\toprule
Parameter & Default & Description \\
\midrule
\texttt{actor\_num\_nodes} & 1 & Nodes for policy training \\
\texttt{actor\_num\_gpus\_per\_node} & 2 & GPUs per node for policy \\
\texttt{vllm\_num\_engines} & 2 & vLLM inference engines \\
\texttt{vllm\_tensor\_parallel\_size} & 1 & Tensor parallelism per engine \\
\texttt{vllm\_gpu\_memory\_utilization} & 0.7 & GPU memory for KV cache \\
\texttt{colocate\_all\_models} & True & Share GPUs across model types \\
\texttt{zero\_stage} & 3 & DeepSpeed ZeRO level \\
\texttt{vllm\_sync\_backend} & nccl & Weight sync backend \\
\bottomrule
\end{tabular}
\end{table}

\subsection{Task-Specific Configuration}

Table~\ref{tab:task_configs} compares the configurations used for MathChat and DSBench experiments.

\begin{table}[h]
\centering
\caption{Configuration differences between MathChat and DSBench.}
\label{tab:task_configs}
\small
\begin{tabular}{@{}lll@{}}
\toprule
Parameter & MathChat & DSBench \\
\midrule
\texttt{rollout\_batch\_size} & 32 & 16 \\
\texttt{n\_samples\_per\_prompt} & 2 & 2 \\
\texttt{num\_episodes} & 8 & 30 \\
\texttt{generate\_max\_len} & 4096 & 16384 \\
\texttt{prompt\_max\_len} & 24576 & 24576 \\
\texttt{coach\_model} & gemini-2.5-flash & gemini-3-pro-preview \\
\texttt{coder\_max\_turns} & 5 & N/A \\
\texttt{filter\_agents\_data} & True & True \\
\texttt{vllm\_gpu\_memory\_utilization} & 0.7 & 0.6 \\
\bottomrule
\end{tabular}
\end{table}

Key differences: DSBench uses longer generation limits (16K vs 4K) to accommodate multi-step data science pipelines, trains for more episodes (30 vs 8) due to smaller dataset size, and uses a more capable coach model (Gemini 3 Pro) for complex ML evaluation. Both use \texttt{filter\_agents\_data=True} to handle variable turn counts across workers.

\subsection{Computational Cost}
\label{app:computational_cost}

\paragraph{Hardware.} All experiments use a single node with 8$\times$ NVIDIA H100 GPUs.

\paragraph{Training Time.} Wall-clock training time is approximately 8--12 hours for 106 training steps on MathChat, with coach API calls being the primary bottleneck (each coach evaluation adds $\sim$2--5 seconds latency). Each rollout generates $\sim$3--9 coach calls (3 agents $\times$ 1--3 turns per agent), totaling $\sim$3,000--10,000 coach calls per training run. DSBench training (30 episodes $\times$ 128 rollouts) incurs similar per-episode costs but with longer per-task execution times due to multi-turn tool use.

\paragraph{API Cost.} Using Gemini 2.5 Flash at approximately \$0.075 per 1M input tokens and \$0.30 per 1M output tokens, the estimated API cost is \$50--150 per training run.

\end{document}